\documentclass[sigconf]{acmart}

\usepackage{booktabs} 
\usepackage{epstopdf} 
\usepackage{verbatim} 
\usepackage{subfigure} 
\usepackage[linesnumbered,ruled]{algorithm2e} 
\usepackage{multirow} 
\usepackage{adjustbox} 
\usepackage{xcolor}

\usepackage{enumitem}

\copyrightyear{2019}
\acmYear{2019} 
\setcopyright{iw3c2w3}
\acmConference[WWW '19]{Proceedings of the 2019 World Wide Web Conference}{May 13--17, 2019}{San Francisco, CA, USA}
\acmBooktitle{Proceedings of the 2019 World Wide Web Conference (WWW '19), May 13--17, 2019, San Francisco, CA, USA}
\acmPrice{}
\acmDOI{10.1145/3308558.3313445}
\acmISBN{978-1-4503-6674-8/19/05}

\usepackage{balance}

\begin{document}
\title{Adversarial Training Methods for Network Embedding}

\author{Quanyu Dai}
\affiliation{%
  \institution{\small The Hong Kong Polytechnic University}
  \city{Kowloon, Hong Kong}
}
\email{dqyzm100@hotmail.com}

\author{Xiao Shen}
\affiliation{%
  \institution{\small The Hong Kong Polytechnic University}
  \city{Kowloon, Hong Kong}
}
\email{shenxiaocam@163.com}

\author{Liang Zhang}
\affiliation{%
  \institution{\small JD.com}
  \city{Beijing, China}
}
\email{zhangliang16@jd.com}

\author{Qiang Li}
\affiliation{%
  \institution{\small Y-tech, Kwai}
  \city{Beijing, China}
}
\email{leetsiang.cloud@gmail.com}

\author{Dan Wang}
\affiliation{%
  \institution{\small The Hong Kong Polytechnic University}
  \city{Kowloon, Hong Kong}
}
\email{csdwang@comp.polyu.edu.hk}

\renewcommand{\shortauthors}{B. Trovato et al.}

\begin{abstract}
  Network Embedding is the task of learning continuous node representations for networks, which has been shown effective in a variety of tasks such as link prediction and node classification. Most of existing works aim to preserve different network structures and properties in low-dimensional embedding vectors, while neglecting the existence of noisy information in many real-world networks and the overfitting issue in the embedding learning process. Most recently, generative adversarial networks (GANs) based regularization methods are exploited to regularize embedding learning process, which can encourage a global smoothness of embedding vectors. These methods have very complicated architecture and suffer from the well-recognized non-convergence problem of GANs. In this paper, we aim to introduce a more succinct and effective local regularization method, namely adversarial training, to network embedding so as to achieve model robustness and better generalization performance. Firstly, the adversarial training method is applied by defining adversarial perturbations in the embedding space with an adaptive $L_2$ norm constraint that depends on the connectivity pattern of node pairs. Though effective as a regularizer, it suffers from the interpretability issue which may hinder its application in certain real-world scenarios. To improve this strategy, we further propose an interpretable adversarial training method by enforcing the reconstruction of the adversarial examples in the discrete graph domain. These two regularization methods can be applied to many existing embedding models, and we take \textsc{DeepWalk} as the base model for illustration in the paper. Empirical evaluations in both link prediction and node classification demonstrate the effectiveness of the proposed methods. The source code is available online\footnote{https://github.com/wonniu/AdvT4NE\_WWW2019}. 
\end{abstract}

%
%

\keywords{Network Embedding, Adversarial Training, Robustness}

\maketitle

\section{Introduction}
 Network embedding strategies, as an effective way for extracting features from graph structured data automatically, have gained increasing attention in both academia and industry in recent years. The learned node representations from embedding methods can be utilized to facilitate a wide range of downstream learning tasks, including some traditional network analysis tasks such as link prediction and node classification, and many important applications in industry such as product recommendation in e-commerce website and advertisement distribution in social networks. Therefore, under such great application interest, substantial efforts have been devoted to designing effective and scalable network embedding models.
 
 Most of the existing works focus on preserving network structures and properties in low-dimensional embedding vectors~\cite{WWW-15-Jian,CIKM-15-SsCao,KDD-16-DxW}. Firstly, \textsc{DeepWalk}~\cite{WWW-15-Jian} defines random walk based neighborhood for capturing node dependencies, and node2vec~\cite{KDD-16-Grover} extends it with more flexibility in balancing local and global structural properties. LINE~\cite{WWW-15-Jian} preserves both first-order and second-order proximities through considering existing connection information. Further, GraRep~\cite{CIKM-15-SsCao} manages to learn different high-order proximities based on different $k$-step transition probability matrix. 
 Aside from the above mentioned structure-preserving methods, several research works investigate the learning of property-aware network embeddings. For example, network transitivity, as the driving force of link formation, is considered in~\cite{KDD-16-Mingdong}, and node popularity, as another important factor affecting link generation, is incorporated into RaRE~\cite{WWW-GuSL018} to learn social-rank aware and proximity-preserving embedding vectors. However, the existence of nosiy information in real-world networks and the overfitting issue in the embedding learning process are neglected in most of these methods, which leaves the necessity and potential improvement space for further exploration.
 
 Most recently, adversarial learning regularization method is exploited for improving model robustness and generalization performance in network embedding~\cite{AAAI-18-Quanyu,KDD-YuZCASZCW18}. ANE~\cite{AAAI-18-Quanyu} is the first try in this direction, which imposes a prior distribution on embedding vectors through adversarial learning. Then, the adversarially regularized autoencoder is adopted in \textsc{NetRA}~\cite{KDD-YuZCASZCW18} to overcome the mode-collapse problem in ANE method. These two methods both encourage the global smoothness of the embedding distribution based on generative adversarial networks (GANs)~\cite{NIPS-14-GoodfellowPMXWOCB}. Thus, they have very complicated frameworks and suffer from the well-recognized hard training problems of GANs~\cite{NIPS-SalimansGZCRCC16,ICML-ArjovskyCB17}. 
 
 In this paper, we aim to leverage the adversarial training (AdvT) method~\cite{ICLR-2014-Szegedy,goodfellow2014explaining} for network embedding to achieve model robustness and better generalization ability. AdvT is a local smoothness regularization method with more succinct architecture. Specifically, it forces the learned classifier to be robust to adversarial examples generated from clean ones with small crafted perturbation~\cite{ICLR-2014-Szegedy}. Such designed noise with respect to each input example is dynamically obtained through finding the direction to maximize model loss based on current model parameters, and can be approximately computed with fast gradient method~\cite{goodfellow2014explaining}. It has been demonstrated to be extremely useful for some classification problems~\cite{goodfellow2014explaining,ICLR-16-Miyato}. 
 
 However, how to adapt AdvT for graph representation learning remains an open problem. It is not clear how to generate adversarial examples in the discrete graph domain since the original method is designed for continuous inputs. In this paper, we propose an adversarial training \textsc{DeepWalk} model, which defines the adversarial examples in the embedding space instead of the original discrete relations and obtains adversarial perturbation with fast gradient method. We also leverage the dependencies among nodes based on connectivity patterns in the graph to design perturbations with different $L_2$ norm constraints, which enables more reasonable adversarial regularization. The training process can be formulated as a two-player game, where the adversarial perturbations are generated to maximize the model loss while the embedding vectors are optimized against such designed noises with stochastic gradient descent method. Although effective as a regularization technique, directly generating adversarial perturbation in embedding space with fast gradient method suffers from interpretability issue, which may restrict its application areas. Further, we manage to restore the interpretability of adversarial examples by constraining the perturbation directions to embedding vectors of other nodes, such that the adversarial examples can be considered as the substitution of nodes in the original discrete graph domain. 
 
 Empirical evaluations show the effectiveness of both adversarial and interpretable adversarial training regularization methods by building network embedding method upon \textsc{DeepWalk}. It is worth mentioning that the proposed regularization methods, as a principle, can also be applied to other embedding models with embedding vectors as model parameters such as node2vec and LINE. The main contributions of this paper can be summarized as follows:
 \begin{itemize}[leftmargin=0.3cm]
 	\item We introduce a novel, succinct and effective regularization technique, namely adversarial training method, for network embedding models which can improve both model robustness and generalization ability.
 	\item We leverage the dependencies among node pairs based on network topology to design perturbations with different $L_2$ norm constraints for different positive target-context pairs, which enables more flexible and effective adversarial training regularization.
 	\item We also equip the adversarial training method with interpretability for discrete graph data by restricting the perturbation directions to embedding vectors of other nodes, while maintaining its usefulness in link prediction and only slightly sacrificing its regularization ability in node classification.
 	\item We conduct extensive experiments to evaluate the effectiveness of the proposed methods. 
 \end{itemize}
 

\section{Background}\label{pre}

 \begin{figure*} \centering
    \subfigure[Cora, training ratio=50\%, 80\%.] { \label{fig:mcc-cora-noise-effect-50}
        \includegraphics[width=0.336\columnwidth]{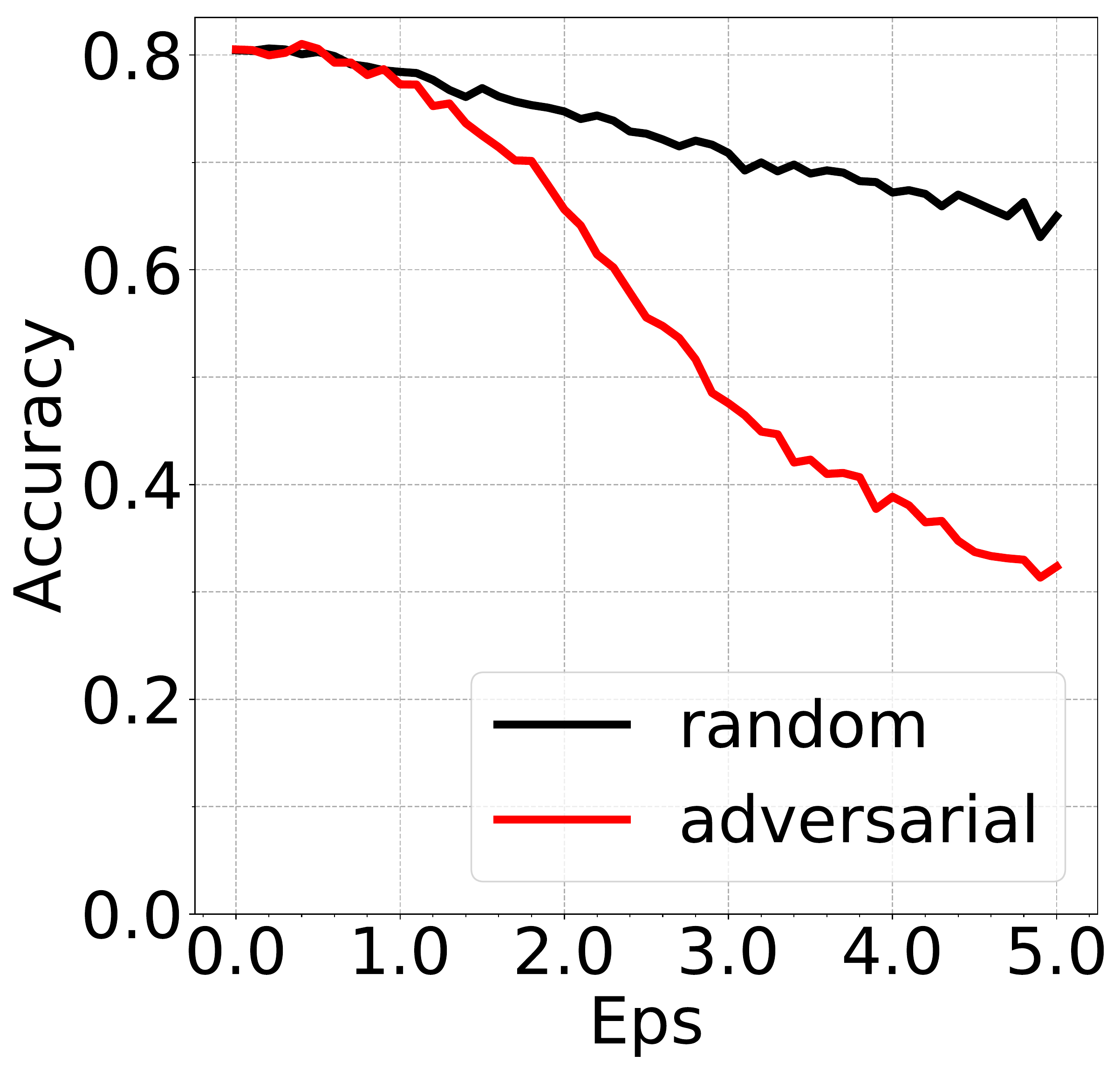}
        \hspace{-0.07in}
        \includegraphics[width=0.336\columnwidth]{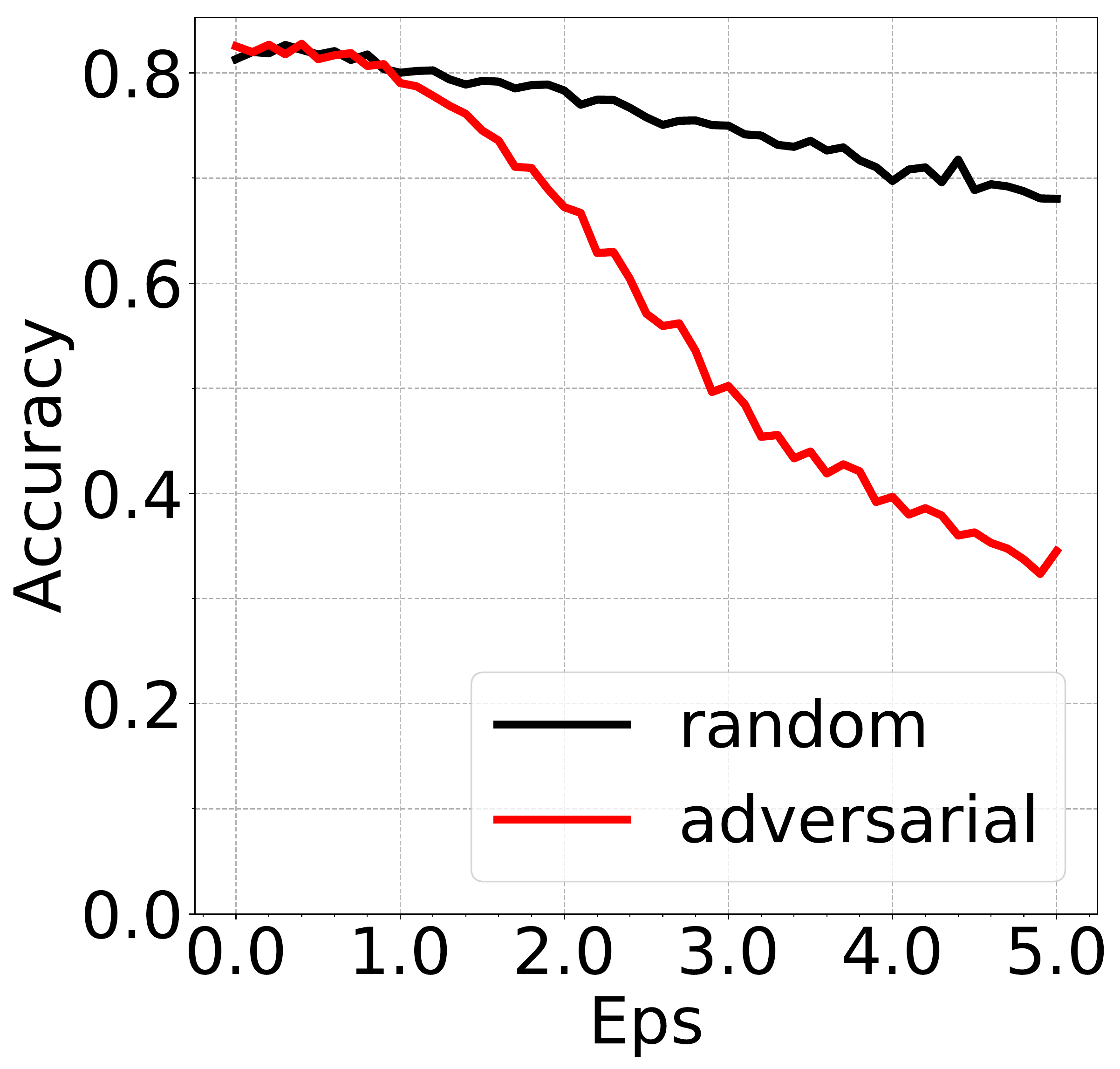}
        }
        \hspace{-0.08in}
    \subfigure[Citeseer, training ratio=50\%, 80\%.] { \label{fig:mcc-citeseer-noise-effect-50}
        \includegraphics[width=0.336\columnwidth]{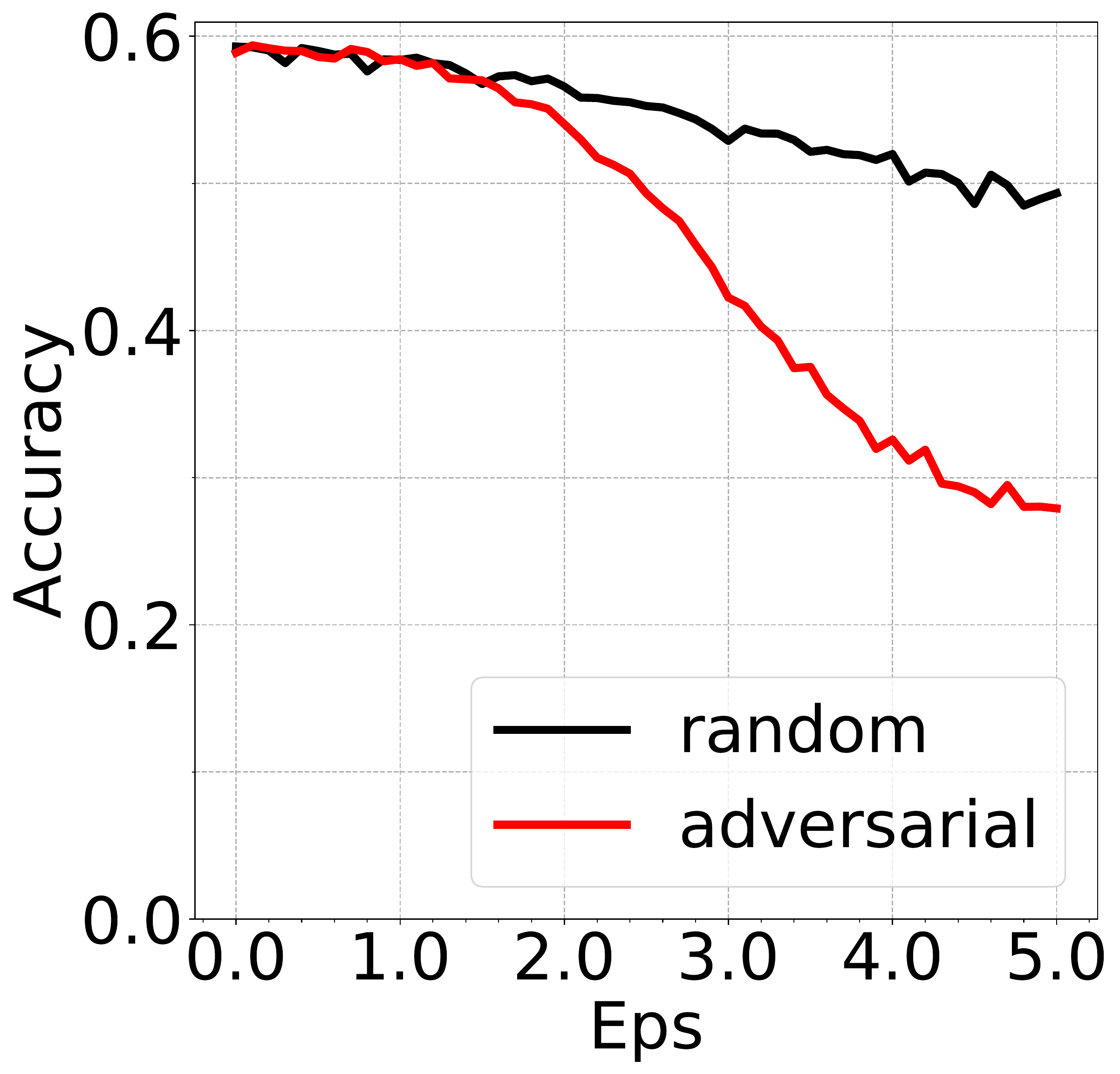}
        \hspace{-0.07in}
        \includegraphics[width=0.336\columnwidth]{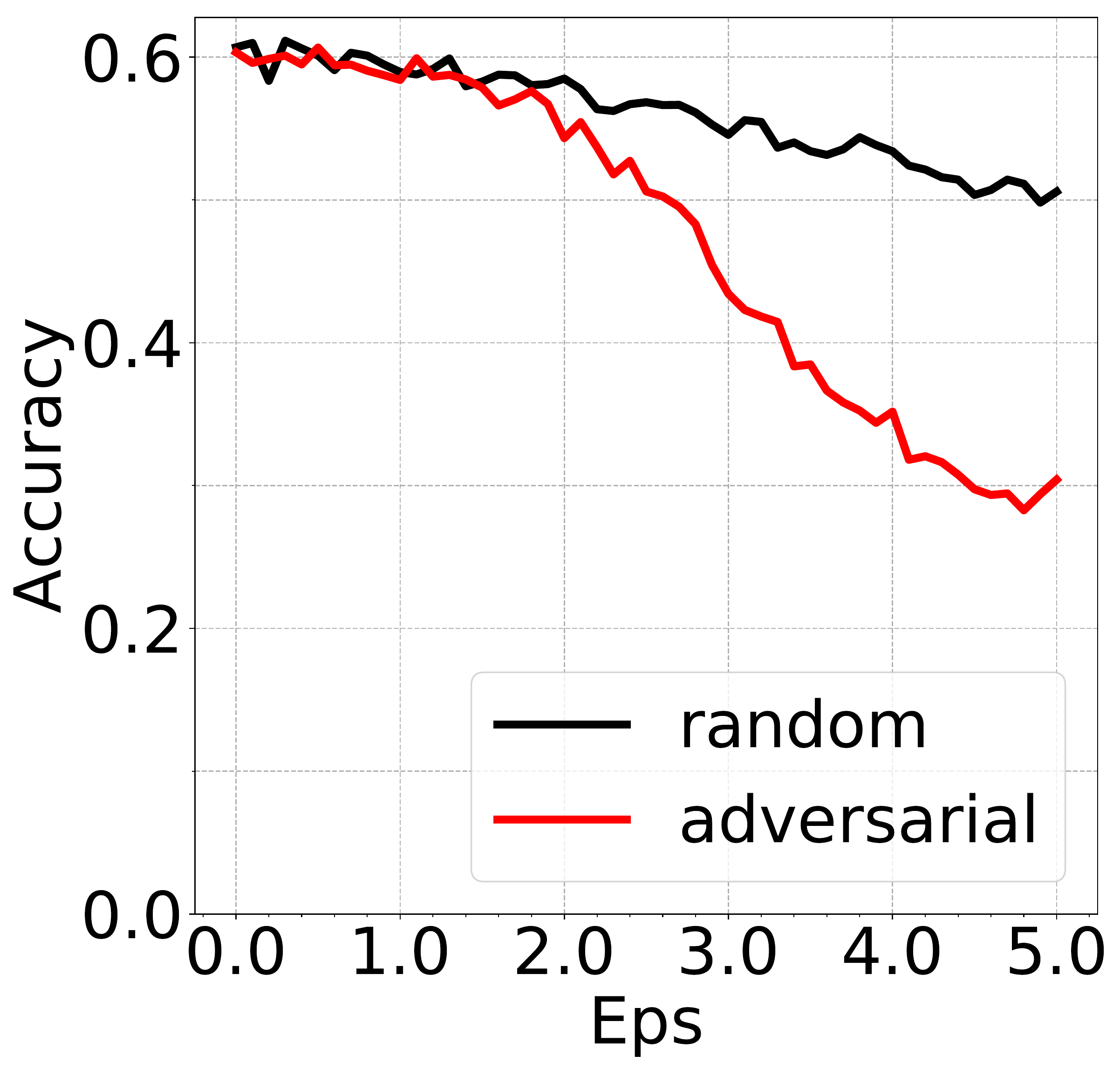}
        }
        \hspace{-0.08in}
    \subfigure[Wiki, training ratio=50\%, 80\%.] { \label{fig:mcc-wiki-noise-effect-50}
        \includegraphics[width=0.336\columnwidth]{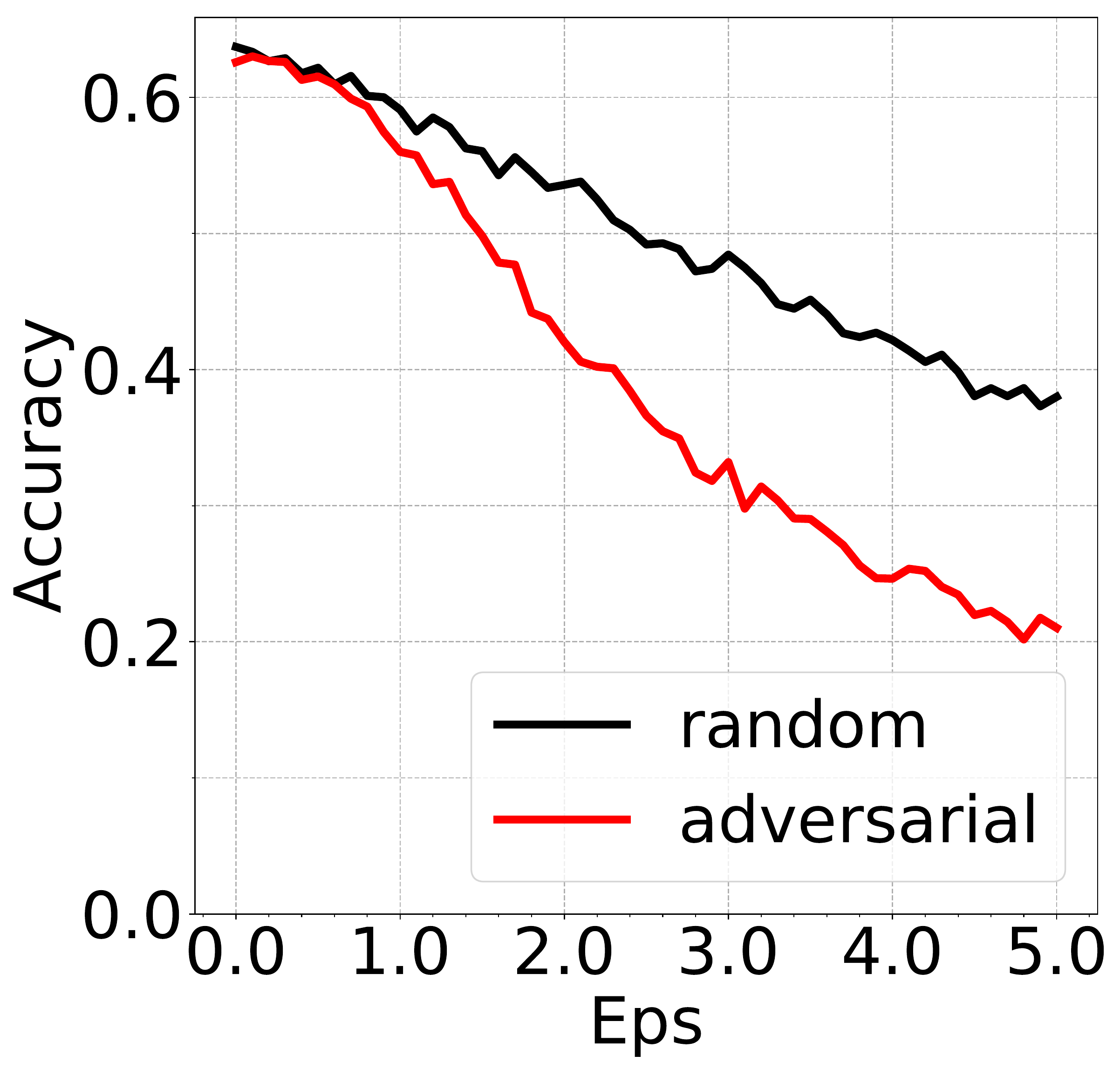}
        \hspace{-0.07in}
        \includegraphics[width=0.336\columnwidth]{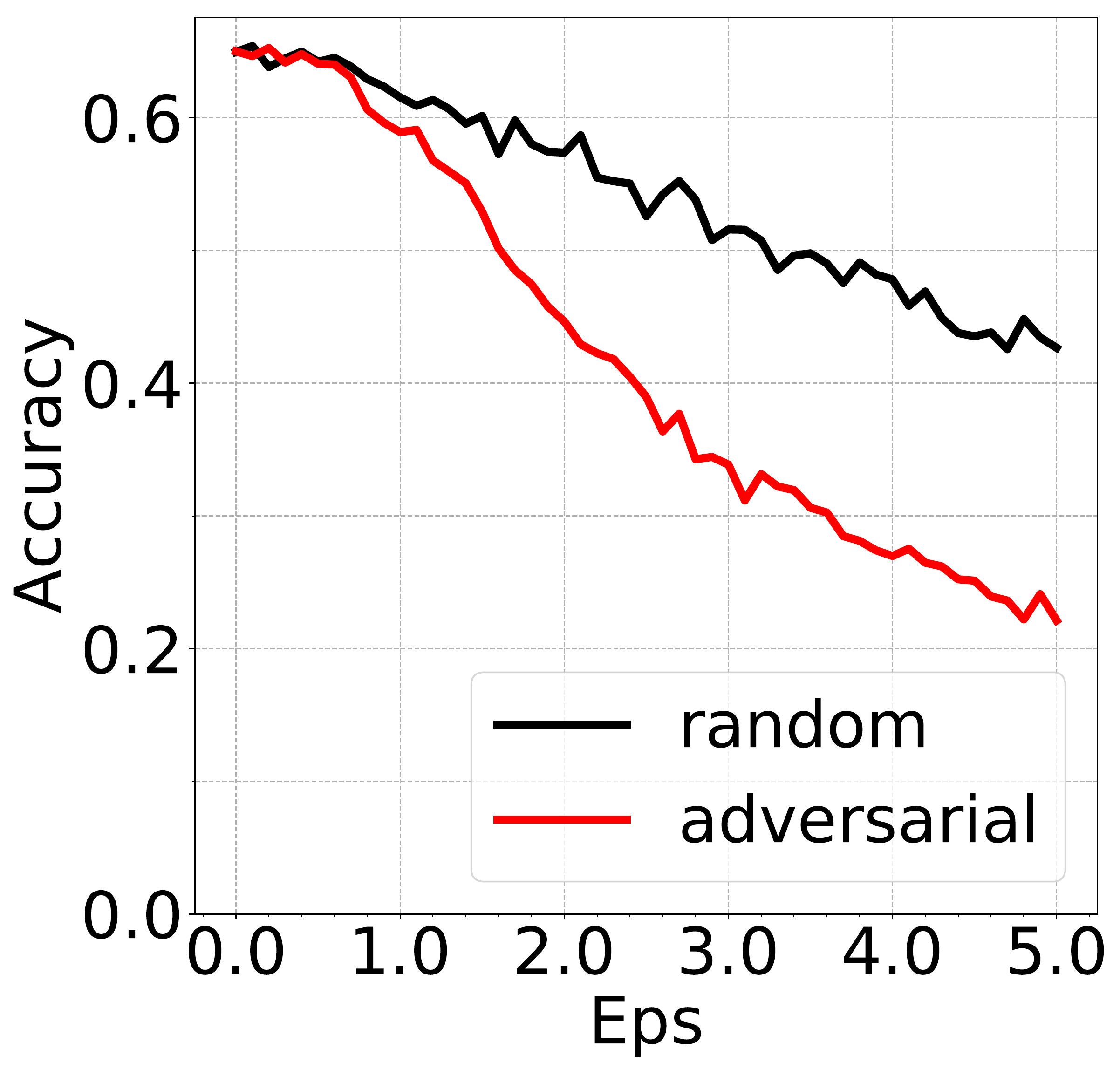}
        }
    \caption{Impact of applying adversarial and random perturbations to the embedding vectors learned by \textsc{DeepWalk} on Cora, Citeseer and Wiki on multi-class classification with training ratio as 50\% and 80\%. Note that "random" represents random perturbations (noises generated from a normal distribution), while "adversarial" represents adversarial perturbations.}
    \label{mcc-noise-effect}
 \end{figure*}

 \subsection{Framework of Network Embedding} \label{pre-tne}
 The purpose of network embedding is to transform discrete network structure information into compact embedding vectors, which can be further used to facilitate downstream learning tasks, such as node classification and link prediction. The research problem can be formally formulated as follows: Given a weighted (unweighted) directed (undirected) graph $G(V, E, A)$ ($N=|V|$), with $V=\{v_i\}_{i=1}^{N}$ as the node set, $E=\{e_{ij}\}_{i,j=1}^N$ as the edge set, and $A$ as the weighted adjacency matrix with $A_{ij}$ quantifying the strength of the relationship between node $v_i$ and $v_j$, network embedding is aimed at learning a mapping function $f: V\mapsto U$, where $U \in R^{N \times d}$ ($d\ll N$) is the embedding matrix with the $ith$ row ${\boldsymbol{u}_i}^T$ as the embedding vector of node $v_i$. Note that for many network embedding models, a context embedding matrix $U^{\prime}$ will also be learned. For these methods, embedding matrix $U$ is also called target embedding matrix.

 The learning framework of many famous network embedding methods, such as \textsc{DeepWalk}~\cite{KDD-14-Bryan}, LINE~\cite{WWW-15-Jian} and node2vec~\cite{KDD-16-Grover}, can be summarized into two phases: a sampling phase that determines node pairs with strong relationships, and an optimization phase that tries to preserve pairwise relationships in the embedding vectors through the negative sampling approach~\cite{NIPS-13-Tomas}. In particular, in the first phase, these three methods capture structural information by defining different neighborhood structures, such as random walk explored neighborhood in~\cite{KDD-14-Bryan,KDD-16-Grover}, first-order and second-order proximities in~\cite{WWW-15-Jian}. We denote the generalized neighbors (not restricted to directly connected nodes) of node $v_i$ as $\mathcal{N}(v_i)$, i.e., nodes in this set are closely related with $v_i$ and should be close with $v_i$ in the embedding space. The loss function of this framework can be abstracted as follows:

 \begin{equation}\label{negative-sampling-loss} \small
 \begin{array}{ll}
 \mathcal{L}(G|\Theta) = - \sum\limits_{v_i\in V}\sum\limits_{v_j \in \mathcal{N}(v_i)} \{ \log\sigma(s(v_i,v_j|\Theta)) \\
 \qquad\qquad \;\;+ \sum\limits_{k=1}^{K}\mathbb{E}_{v_k\sim P_k(v)}[\log\sigma(-s(v_i,v_k|\Theta))] \},
 \end{array}
 \end{equation}
 where $\Theta$ represents model parameters such as target and context embedding matrices, $s(v_i,v_j|\Theta)$ represents the similarity score of node $v_i$ and $v_j$ based on model parameters $\Theta$, and $\sigma(\cdot)$ is the sigmoid function. $P_k(v)$ denotes the distribution for sampling negative nodes, and a simple variant of unigram distribution is usually utilized, i.e., $P_k(v)\propto d_v^{3/4}$, where $d_v$ is the out-degree of node $v$. Eq.~(\ref{negative-sampling-loss}) is actually a cross entropy loss with closely related node pair $(v_i, v_j)$ as positive samples and $(v_i,v_k)$ as negative samples, and thus network embedding can be considered as a classification problem. 

 \subsection{Adversarial Training}
 Adversarial training~\cite{ICLR-2014-Szegedy,goodfellow2014explaining} is a newly proposed effective regularization method for classifiers which can not only improve the robustness of the model against adversarial attacks, but also achieve better generalization performance on learning tasks. It augments the original clean data with the dynamically generated adversarial examples, and then trains the model with the newly mixed examples. Denote the input as $\boldsymbol{x}$ and model parameters as $\boldsymbol{\theta}$. The loss on adversarial examples can be considered as a regularization term in the trained classifier $p(y|\cdot)$, which is as follows:
 \begin{equation}\label{adv-regularization}\small
 - \log p(y|\boldsymbol{x} + \boldsymbol{n}_{adv};\boldsymbol{\theta}),
 \end{equation}
 \begin{equation}\label{adv-generation}\small
 where \; \boldsymbol{n}_{adv}=\mathop{\arg\min}_{\boldsymbol{n},\, \|\boldsymbol{n}\|\leq\epsilon}\log p(y|\boldsymbol{x} + \boldsymbol{n};\hat{\boldsymbol{\theta}}),
 \end{equation}
 where $\boldsymbol{n}$ is the perturbation on the input, $\epsilon$ represents the norm constraint of $\boldsymbol{n}$, and $\hat{\boldsymbol{\theta}}$ are current model parameters but fixed as constants. We employ $L_2$ norm in this paper, while $L_1$ norm has also been used in the literature~\cite{goodfellow2014explaining}. Eq.~(\ref{adv-regularization}) means that the model should be robust on the adversarial perturbed examples. Before each batch training, the adversarial noise $\boldsymbol{n}$ with respect to the input $\boldsymbol{x}$ is firstly generated by solving optimization problem~(\ref{adv-generation}) to make it resistant to current model. Since it is difficult to calculate Eq.(~\ref{adv-generation}) exactly in general, fast gradient descent method~\cite{goodfellow2014explaining} is widely used to obtain the adversarial noise approximately by linearizing $\log p(y|\boldsymbol{x};\boldsymbol{\theta})$ around $\boldsymbol{x}$. Specifically, the adversarial perturbation with $L_2$ norm constraint can be obtained as follows:
 \begin{equation}\label{fast-gradient}\small
 \boldsymbol{n}_{adv} = - \epsilon\cdot\frac{\boldsymbol{g}}{\|\boldsymbol{g}\|_2}\;where\; \boldsymbol{g}=\nabla_{\boldsymbol{x}}\log p(y|\boldsymbol{x};\hat{\boldsymbol{\theta}}).
 \end{equation}
 It can be easily calculated with backpropagation method.

 \subsection{Motivation}

 To improve the generalization ability of network embedding models, two ways have been used: firstly, some denoising autoencoder based methods~\cite{AAAI-16-SsCao,AAAI-18-Quanyu} improve model robustness by adding random perturbation to input data or hidden layers of deep models; secondly, some existing methods~\cite{AAAI-18-Quanyu,KDD-YuZCASZCW18} regularize embedding vectors from a global perspective through GAN-based method, i.e., encouraging the global smoothness of the distribution of embeddings. In this paper, we aim to introduce a novel, more succinct and effective regularization method for network embedding models, i.e., adversarial training (AdvT)~\cite{goodfellow2014explaining}. AdvT generates crafted adversarial perturbations to model inputs and encourages local smoothness for improving model robustness and generalization performance, which can be expected to be more effective than the random perturbation methods~\cite{AAAI-16-SsCao} and global regularization methods~\cite{AAAI-18-Quanyu,KDD-YuZCASZCW18}. In the following, we would like to compare the impact of adversarial and random perturbation on embedding vectors to better motivate this new regularization method.

 However, it is not clear how to integrate adversarial training into existing network embedding methods. Graph data is discrete, and the continuous adversarial noise can not be directly imposed on the discrete connected information. To bypass this difficulty, we seek to define the adversarial perturbation on embedding vectors instead of the discrete graph domain as inspired by~\cite{ICLR-16-Miyato}. We define the adversarial perturbation on node embeddings as follows:
 \begin{equation}\small
 \boldsymbol{n}_{adv} = \mathop{\arg\max}_{\boldsymbol{n},\, \|\boldsymbol{n}\|\leq\epsilon} \mathcal{L}(G|\hat{\Theta}+\boldsymbol{n}),
 \end{equation}
 which can be further approximated with fast gradient method as presented in Eq.~(\ref{fast-gradient}).

 Take \textsc{DeepWalk}~\cite{KDD-14-Bryan} with negative sampling loss as an illustrative example. We explore the effect of adversarial perturbations on embedding vectors by adding them to the learned embedding vectors from \textsc{DeepWalk}, and then perform multi-class classification with the perturbed embeddings on several datasets. Besides, we choose random perturbations as the compared baseline, i.e.,  noises generated from a normal distribution. Figure~\ref{mcc-noise-effect} displays node classification results with varying $L_2$ norm constraints on the perturbations. We can find that embedding vectors are much more vulnerable to adversarial perturbations than random ones. For example, when $\varepsilon$ is set to 2.0, the performance of node classification with training ratio as 80\% on Cora drops 3.35\% under random perturbation, while that decreases 16.25\% under adversarial perturbation which is around 4 times more serious. If the embedding vectors can be trained to be more robust on adversarial noises, we can expect more significant improvements in generalization performance.

\section{Proposed Methods}
 \begin{figure*}[t]
    \centering
    \includegraphics[width=1.9\columnwidth]{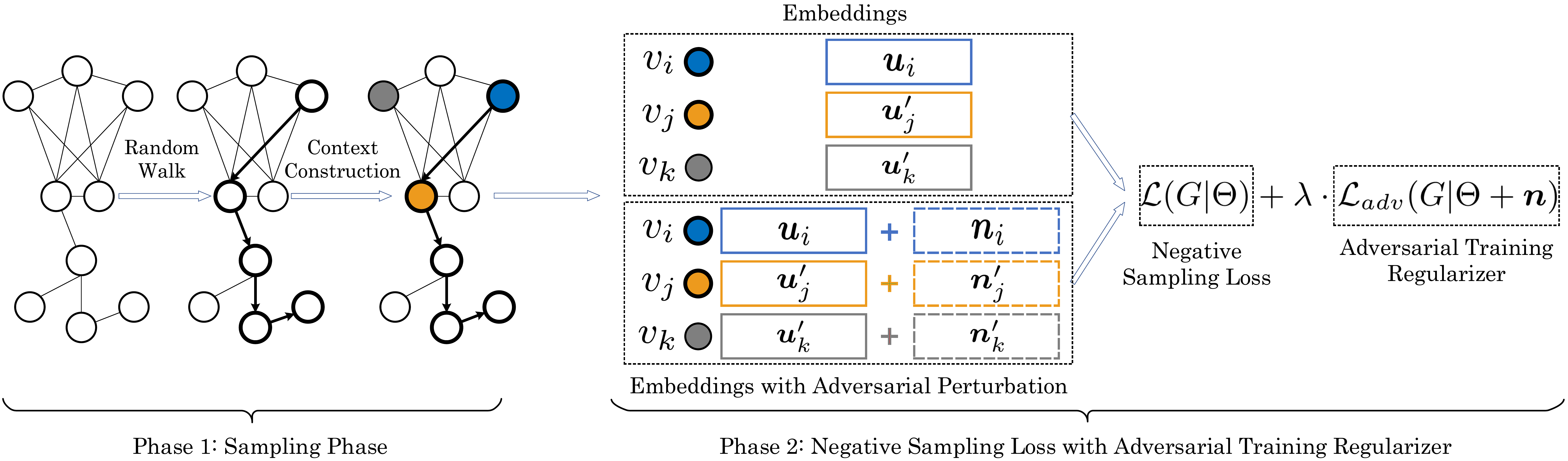}
    \caption{\textsc{DeepWalk} with Adversarial Training Regularization}
    \label{ADW-Framework}
 \end{figure*}

 In this section, we first describe the adapted adversarial training method for network embedding models, and present the algorithm based on \textsc{DeepWalk}. Then, we will tackle its interpretability issue by designing a new adversarial perturbation generation method.

 \subsection{Adversarial Training \textsc{DeepWalk}}
 Figure~\ref{ADW-Framework} shows the framework of \textsc{DeepWalk} with adversarial training regularization. It consists of two phases: a sampling phase that determines node pairs with strong relationships, and an optimization phase that tries to preserve pairwise relationships in the embedding vectors based on negative sampling approach. Note that in this paper we take \textsc{DeepWalk} as the illustrative example, and the proposed framework can be applied to the network embedding methods, such as LINE and node2vec, with the main difference in the sampling phase only.

 In the first phase, \textsc{DeepWalk} transforms the network into node sequences by truncated random walk. For each node $v_i\in V$, $\eta$ sequences each with $l$ nodes will be randomly sampled based on network structure with $v_i$ as the starting point. In every walking step, the next node $v_j$ will be sampled from the neighbors of current node $v_k$ with the probability proportional to the edge strength $A_{kj}$ between $v_k$ and $v_j$. In practice, the alias table method~\cite{KDD-LiARS14} is usually leveraged for node sampling given the weight distribution of neighbors of current node, which only takes $O(1)$ time in a single sampling step. Then in the context construction process, closely related node pairs will be determined based on the sampled node sequences. Denote a node sequence as $S$ with the $ith$ node as $s_i$. The positive target-context pairs from $S$ is defined as $\mathcal{P} = \{(s_i, s_j): |i-j|\leq c\}$, where $c$ represents the window size. With the constructed node pairs, the negative sampling loss will be optimized, which is defined as follows:
 \begin{equation}\label{dw-loss}\small
 \begin{array}{ll}
 \mathcal{L}(G|\Theta) = - \sum\limits_{v_i\in V}\sum\limits_{v_j \in \mathcal{N}(v_i)} \{ \log\sigma({\boldsymbol{u}_j^{\prime}}^T\cdot \boldsymbol{u}_i) \\
 \qquad\qquad \;\; + \sum\limits_{k=1}^{K}\mathbb{E}_{v_k\sim P_k(v)}[\log\sigma(-{\boldsymbol{u}_k^{\prime}}^T\cdot \boldsymbol{u}_i)] \},
 \end{array}
 \end{equation}
 where $(v_i, v_j)$ is from the constructed positive target-context pairs, and $\boldsymbol{u}_i$ and $\boldsymbol{u}_j^{\prime}$ are the target embedding of node $v_i$ and context embedding of node $v_j$ respectively.

 For the adversarial version of \textsc{DeepWalk}, an adversarial training regularization term is added to the original loss to help learn robust node representations against adversarial perturbations. The regularization term shares the same set of model parameters with the original model, but with the perturbed target and context embeddings as input. Existing methods consider the input examples independently, and impose the unique $L_2$ norm constraint on all adversarial perturbations~\cite{goodfellow2014explaining,ICLR-16-Miyato}. For graph structured data, entities often correlate with each other in a very complicated way, so it is inappropriate to treat all positive target-context relations equally without discrimination. Adversarial regularization helps alleviate overfitting issue, but it may also bring in some noises that can hinder the preservation of structural proximities, i.e., adding noises to those inherently closely-related node pairs will prevent them from having similar embeddings. Thus, we take advantages of the denpendencies among nodes to assign different $L_2$ norm constraints to different positive target-context relations adaptively. Specifically, the more closely two nodes are connected, the smaller the constraint should be. The intuition is that less noises should be added to those node pairs which are inherently strongly-connected in the original network, thus they can be pushed closer in the embedding space with high flexibility, while for those weakly-connected pairs larger constraint can help alleviate the overfitting issue. 
 
 We obtain the similarity score of two nodes through computing the shifted positive pointwise mutual information matrix~\cite{NIPS-14-LevyG}:
 \begin{equation}\small
 M_{ij} = \max\{\log(\frac{\hat{M}_{ij}}{\sum_{k}\hat{M}_{kj}})-\log(\beta), 0\},
 \end{equation}
 where $\hat{M}=\hat A+\hat A^2+\cdots +\hat A^t$ captures different high-order proximities, $\hat{A}$ is the $1$-step probability transition matrix obtained from $A$ after the row-wise normalization, and $\beta$ is a shift factor. We set $t$ to 2 and $\beta$ to $\frac{1}{N}$ in the experiments. Then, the adaptive scale factor for the $L_2$ norm constraint of the target-context pair $v_i$ and $v_j$ is calculated as below:
 \begin{equation}\small
 \Phi_{ij} = 1 - M_{ij}/\max\{M\},
 \end{equation}
 where $\max\{M\}$ represents the maximum entry of matrix $M$. Since $M_{ij}>0 \;(\forall i, j)$, $\Phi_{ij}\in [0, 1]$. For those strongly-connected target-context pairs, the adaptive scale factor can help scale down the $L_2$ norm constraint of the adversarial perturbation, and thus alleviate the negative effect from the noises.
 
 Then, the adversarial training regularizer with scale factor for $L_2$ norm constraint is defined as follows:
 \begin{equation} \label{adv-loss}\small
 \begin{array}{ll}
 \mathcal{L}_{adv}(G|\Theta+\boldsymbol{n}_{adv})\\
 = - \sum\limits_{v_i\in V}\sum\limits_{v_j \in \mathcal{N}(v_i)} \{ \log\sigma((\boldsymbol{u}_j^{\prime}+ \Phi_{ij}\cdot(\boldsymbol{n}_j^{\prime})_{adv})^T\cdot (\boldsymbol{u}_i+\Phi_{ij}\cdot(\boldsymbol{n}_i)_{adv})) \\
 \quad+ \sum\limits_{k=1}^{K}\mathbb{E}_{v_k\sim P_k(v)}[\log\sigma(-(\boldsymbol{u}_k^{\prime}+(\boldsymbol{n}_k^{\prime})_{adv})^T\cdot (\boldsymbol{u}_i+(\boldsymbol{n}_i)_{adv}))] \},
 \end{array}
 \end{equation}
 where $(\boldsymbol{n}_i)_{adv}$ and $(\boldsymbol{n}_j^{\prime})_{adv}$ represent the original adversarial perturbation for target embedding of node $v_i$ and context embedding of node $v_j$ respectively.

 Finally, one key problem is how to compute the adversarial perturbation for the given embedding vector of a node $v$. Here, we follow the famous adversarial training method directly~\cite{ICLR-2014-Szegedy,goodfellow2014explaining}, and generate the perturbation noises to maximize model loss under current model parameters. The adversarial perturbation for node $v$ is defined as follows:
 \begin{equation}\small
 \boldsymbol{n}_{adv} = \mathop{\arg\max}_{\boldsymbol{n},\, \|\boldsymbol{n}\|\leq\epsilon} \mathcal{L}(G|\hat{\Theta}+\boldsymbol{n}).
 \end{equation}
 It can be further approximated with fast gradient method as follows:
 \begin{equation}
 \boldsymbol{n}_{adv} =  \epsilon\frac{\boldsymbol{g}}{\|\boldsymbol{g}\|_2}\;where\; \boldsymbol{g}=\nabla_{\boldsymbol{u}}\mathcal{L}(G|\hat{\Theta}).
 \end{equation}

 Therefore, the overall loss for the proposed adversarial training \textsc{DeepWalk} is defined as follows:
 \begin{equation}\label{adw-loss}\small
 \mathcal{L}(G|\Theta) + \lambda \cdot \mathcal{L}_{adv}(G|\Theta+\boldsymbol{n}_{adv}),
 \end{equation}
 where $\lambda$ is a hyperparameter to control the importance of the regularization term.

 In this paper, we utilize \textsc{DeepWalk} with negative sampling loss as the base model for building the adversarial version of network embedding methods. Since the original implementation is based on the well encapsulated library, which lacks flexibility for further adaption, we re-implement the model with tensorflow~\cite{OSDI-16-AbadiBCCDDDGIIK16} and utilize a slightly different training strategy. Specifically, in each training epoch, we independently construct positive target-context pairs with random walk based method, and then optimize model parameters with mini-batch stochastic gradient descent technique. Algorithm~\ref{adw-algorithm} summarizes the training procedure for the adversarial training \textsc{DeepWalk}. The model parameters are firstly initialized by training \textsc{DeepWalk} with the method introduced above. For each batch training, adversarial perturbations are generated with fast gradient method for each node in the batch as presented in Line $7$. Then, the target and context embeddings will be updated by optimizing the negative sampling loss with adversarial training regularization as shown in Line $9$. Asynchronous version of stochastic gradient descent~\cite{NIPS-11-FengNiu} can be utilized to accelerate the training as \textsc{DeepWalk}. Note that we ignore the derivative of $\boldsymbol{n}_{adv}$ with respect to model parameters. The adversarial perturbations can be computed with simple back-propagation method, which enjoys low computational cost. Thus, the adversarial training \textsc{DeepWalk} is scalable as the base model.

 \begin{algorithm}[t]\small
 \caption{The adversarial training \textsc{DeepWalk}}
 \label{adw-algorithm}
    \SetKwInOut{Input}{Input}
    \SetKwInOut{Output}{Output}

    \Input{graph $G(V, E, A)$, window size $c$, embedding size $d$, walks per node $\eta$, negative size $K$, walk length $l$, adversarial noise level $\epsilon$, adversarial regularization strength $\lambda$, batch size $b$}
    \Output{Embedding matrix $U$}
    Initialize target and context embeddings with \textsc{DeepWalk}\;
    \While{not converge}{
        Generate a set of positive target-context pairs $\mathcal{P}$ with random walk based method\;
        \Repeat{$[|\mathcal{P}|/b]_+$ times}{
         Sample a batch $\mathcal{B}$ of target-context pairs from $\mathcal{P}$\;
         $//$ Generate adversarial perturbations \\
         $\boldsymbol{n}_{adv} = \epsilon \frac{\boldsymbol{g}}{\|\boldsymbol{g}\|_2}$ where $ \boldsymbol{g}=\nabla_{\boldsymbol{u}}\mathcal{L}(G|\hat{\Theta})$ for each node $v$ in the batch\;
         $//$ Optimize model parameters \\
         Update target and context embeddings by applying gradient descent technique to Eq.~(\ref{adw-loss}) \;
        }
    }
 \end{algorithm}

 \subsection{Interpretable Adversarial Training \textsc{DeepWalk}}
 Adversarial examples refer to examples that are generated by adding viciously designed perturbations with norm constraint to the clean input, which can significantly increase model loss and probably induce prediction error~\cite{ICLR-2014-Szegedy}. Take an example from~\cite{goodfellow2014explaining} for illustration, a ``panda" image with imperceptibly small adversarial perturbation is assigned to be ``gibbon" by the well-trained classification model with high confidence, while the original image can be correctly classified. Such adversarial examples can be well interpreted since the perturbations are imposed on the input space. For the adversarial training \textsc{DeepWalk}, adversarial perturbations are added to node embeddings instead of the discrete nodes and connected relations, and thus can not be easily reconstructed in the discrete graph domain. Though effective as a regularizer for improving model generalization performance, it suffers from lack of interpretability, which may create a barrier for its adoption in some real-world applications.

 In this section, we propose an interpretable adversarial \textsc{DeepWalk} model by restoring the interpretability of adversarial perturbations. Instead of pursuing the worst perturbation direction only, we restrict the direction of perturbations toward a subset of nodes in the graph in the embedding space, such as the neighbors of the considered node. In this way, the adversarial perturbations in the node embedding space might be interpreted as the substitution of nodes in the original input space, i.e., the discrete target-context relations. However, there might be a certain level of sacrifice on the regularization performance because of the restriction on perturbation directions.

 The \textit{direction vector} from node $v_t$ to $v_k$ in the embedding space is defined as follows:
 \begin{equation}\small
 \boldsymbol{v}^{(t)}_{k} = \frac{\tilde{\boldsymbol{v}}^{(t)}_{k}}{\|\tilde{\boldsymbol{v}}^{(t)}_{k}\|_2}, \; where\; \tilde{\boldsymbol{v}}^{(t)}_{k}=\boldsymbol{u}_k-\boldsymbol{u}_t.
 \end{equation}
 Denote $V^{(t)} \subseteq V$ ($|V^{(t)}|=T$, $|V^{(t)}|\ll |V|$) as a set of nodes for generating adversarial perturbation for node $v_t$. We define $V^{(t)}$ as the top $T$ nearest neighbors of node $v_t$ in the embedding space based on current model parameters. To improve model efficiency, we can also obtain $V^{(t)}$ based on the pretrained model parameters, and fix it for all training epochs. We use the latter strategy for experiments in this paper. Denote $\boldsymbol{w}^{(t)}\in \mathcal{R}^T$ as the weight vector for node $v_t$ with $w^{(t)}_k$ representing the weight associated with direction vector $\boldsymbol{v}^{(t)}_{k}$, where $v_k$ is the $kth$ node in $V^{(t)}$. The interpretable perturbation for $v_t$ is defined as the weighted sum of the direction vectors starting from $v_t$ and ending with nodes in $V^{(t)}$:
 \begin{equation}\label{iadv-example}\small
 \boldsymbol{n}(\boldsymbol{w}^{(t)}) = \sum\limits_{k=1}^{T} w^{(t)}_k \boldsymbol{v}^{(t)}_{k},\;v_k \in V^{(t)},\;\forall k=1,\cdots,T.
 \end{equation}
 The adversarial perturbation is obtained by finding the weights that can maximize the model loss:
 \begin{equation}\small
 \boldsymbol{w}_{iAdv}^{(t)} = \mathop{\arg\max}_{\boldsymbol{w}^{(t)},\,\|\boldsymbol{w}^{(t)}\|\leq \epsilon} \mathcal{L}_{iAdv}(G|\Theta+\boldsymbol{n}(\boldsymbol{w}^{(t)})),
 \end{equation}
 where $\mathcal{L}_{iAdv}$ is obtained by replacing $\boldsymbol{n}_{adv}$ in Eq.~(\ref{adv-loss}) with $\boldsymbol{n}(\boldsymbol{w}^{(t)})$. In consideration of model efficiency, the above regularization term is approximated with first-order Taylor series for easy computation as in~\cite{goodfellow2014explaining}. Thus, the weights for constructing interpretable adversarial perturbation for node $v_t$ can be computed as follows:
 \begin{equation}\small
 \boldsymbol{w}_{iAdv}^{(t)} = \epsilon \frac{\boldsymbol{g}}{\|\boldsymbol{g}\|_2}\;where\; \boldsymbol{g}=\nabla_{\boldsymbol{w}^{(t)}}\mathcal{L}_{iAdv}(G|\Theta+\boldsymbol{n}(\boldsymbol{w}^{(t)})).
 \end{equation}
 Substituting $\boldsymbol{w}_{iAdv}^{(t)}$ into Eq.~(\ref{iadv-example}), we can get the adversarial perturbations $\boldsymbol{n}(\boldsymbol{w}^{(t)}_{iAdv})$. Further, by replacing $\boldsymbol{n}_{adv}$ with $\boldsymbol{n}(\boldsymbol{w}^{(t)}_{iAdv})$ in Eq.~(\ref{adv-loss}), we can have the interpretable adversarial training regularizer for \textsc{DeepWalk}. The algorithm for interpretable adversarial training \textsc{DeepWalk} is different from Algorithm~\ref{adw-algorithm} in the way of generating adversarial perturbations, and thus we do not present it in this paper due to space limitation. Since $|V^{(t)}|\ll |V|$, the computation of adversarial perturbation for one node takes constant time. Therefore, the time complexity of this model is also linear to the number of nodes in the graph as \textsc{DeepWalk}.

\section{Experiments} \label{experiments}
 
 In this section, we empirically evaluate the proposed methods through performing link prediction and node classification on several benchmark datasets.

 \begin{table}[t]
 \centering
 \caption{Statistics of benchmark datasets}
 \begin{adjustbox}{max width=0.40\textwidth}
 \begin{tabular}{ c | c | c | c | c | c }
 \hline
 \hline
 Name & Cora & Citeseer & Wiki & CA-GrQc & CA-HepTh \\
 \hline
 $|V|$ & 2,708 & 3,264 & 2,363 & 5,242 & 9,877 \\
 $|E|$ & 5,278 & 4,551 & 11,596 & 14,484 & 25,973  \\
 Avg. degree & 1.95 & 1.39 & 4.91 & 2.76 & 2.63 \\
 \#Labels & 7 & 6 & 17 & - & - \\
 \hline
 \hline
 \end{tabular}
 \end{adjustbox}
 \label{tab-dataset}
 \end{table}

\subsection{Experiment Setup}
\subsubsection{Datasets}
 We conduct experiments on several benchmark datasets from various real-world applications. Table~\ref{tab-dataset} shows some statistics of them. Note that we do some preprocessing on the original datasets by deleting self-loops and nodes with zero degree. Some descriptions of these datasets are summarized as follows:

 \begin{itemize}[leftmargin=0.3cm]
  \item \textbf{Cora, Citeseer}~\cite{Retr-00-McCallumNRS}: Paper citation networks. Cora consists of 2708 papers with 7 categories, and Citeseer consists 3264 papers including 6 categories.
  \item \textbf{Wiki}~\cite{AI-M-08-Sen}: Wiki is a network with nodes as web pages and edges as the hyperlinks between web pages.
  \item \textbf{CA-GrQc, CA-HepTh}~\cite{TKDD-07-LeskovecKF}: Author collaboration networks. They describe scientific collaborations between authors with papers submitted to General Relativity and Quantum Cosmology category, and High Energy Physics, respectively. 
  \end{itemize}

\subsubsection{Baseline Models}
 The descriptions of the baseline models are as follows:

 \begin{itemize}[leftmargin=0.3cm]
    \item \textbf{Graph Factorization (GF)}~\cite{WWW-AhmedSNJS13}: GF directly factorizes the adjacency matrix with stochastic gradient descent technique to obtain the embeddings, which enables it scale to large networks.
    \item \textbf{\textsc{DeepWalk}}~\cite{KDD-14-Bryan}: \textsc{DeepWalk} regards node sequence obtained from truncated random walk as word sequence, and then uses skip-gram model to learn node representations. 
    We directly use the publicly available source code with hierarchical softmax approximation for experiments.
    \item \textbf{LINE}~\cite{WWW-15-Jian}: LINE preserves network structural proximities through modeling node co-occurrence probability and node conditional probability, and leverages the negative sampling approach to alleviate the expensive computation.
    \item \textbf{node2vec}~\cite{KDD-16-Grover}: node2vec differs from \textsc{DeepWalk} by proposing more flexible method for sampling node sequences to strike a balance between local and global structural properties. 
    \item \textbf{GraRep}~\cite{CIKM-15-SsCao}: GraRep applies SVD technique to different $k$-step probability transition matrix to learn node embeddings, and finally obtains global representations through concatenating all $k$-step representations.
    \item \textbf{AIDW}~\cite{AAAI-18-Quanyu}: AIDW is an inductive version of \textsc{DeepWalk} with GAN-based regularization method. A prior distribution is imposed on node representations through adversarial learning to achieve a global smoothness in the distribution.
 \end{itemize}

 Our implemented version of \textsc{DeepWalk} is based on negative sampling approach, thus we denote it as \textsc{Dwns} to avoid confusion. We also include a baseline, namely \textsc{Dwns}\_rand, with noises from a normal distribution as perturbations in the regularization term. Following existing work~\cite{IJCAI-SatoSS018}, we denote the adversarial training \textsc{DeepWalk} as \textsc{Dwns}\_AdvT, and the interpretable adversarial training \textsc{DeepWalk} as \textsc{Dwns}\_iAdvT in the rest of the paper. 
 
\subsubsection{Parameter Settings}
 
 For \textsc{Dwns} and its variants including \textsc{Dwns}\_rand, \textsc{Dwns}\_AdvT and \textsc{Dwns}\_iAdvT, the walk length, walks per node, window size, negative size, regularization strength, batch size and learning rate are set to 40, 1, 5, 5, 1, 1024 and 0.001, respectively. The adversarial noise level $\epsilon$ has different settings in \textsc{Dwns}\_AdvT and \textsc{Dwns}\_iAdvT, while \textsc{Dwns}\_rand follows the settings of \textsc{Dwns}\_AdvT. For \textsc{Dwns}\_AdvT, $\epsilon$ is set to different value for different datasets. 
 Specifically, $\epsilon$ is set to 0.9 for Cora, 1.1 for Citeseer in both link prediction and node classification, and 0.6 and 0.5 for Wiki in node classification and link prediction respectively, while it is set to 0.5 for all other datasets in these two learning tasks. For \textsc{Dwns}\_iAdvT, $\epsilon$ is set to 5 for all datasets in both node classification and link prediction tasks, and the size of the nearest neighbor set $T$ is set to 5. Besides, the dimension of embedding vectors are set to 128 for all methods. 

\subsection{Impact of Adversarial Training Regularization}\label{sec-adv-effect}

\begin{figure*} \centering
	\subfigure[Cora.] { \label{fig:mcc-cora-adv-effect}
		\includegraphics[width=0.338\columnwidth]{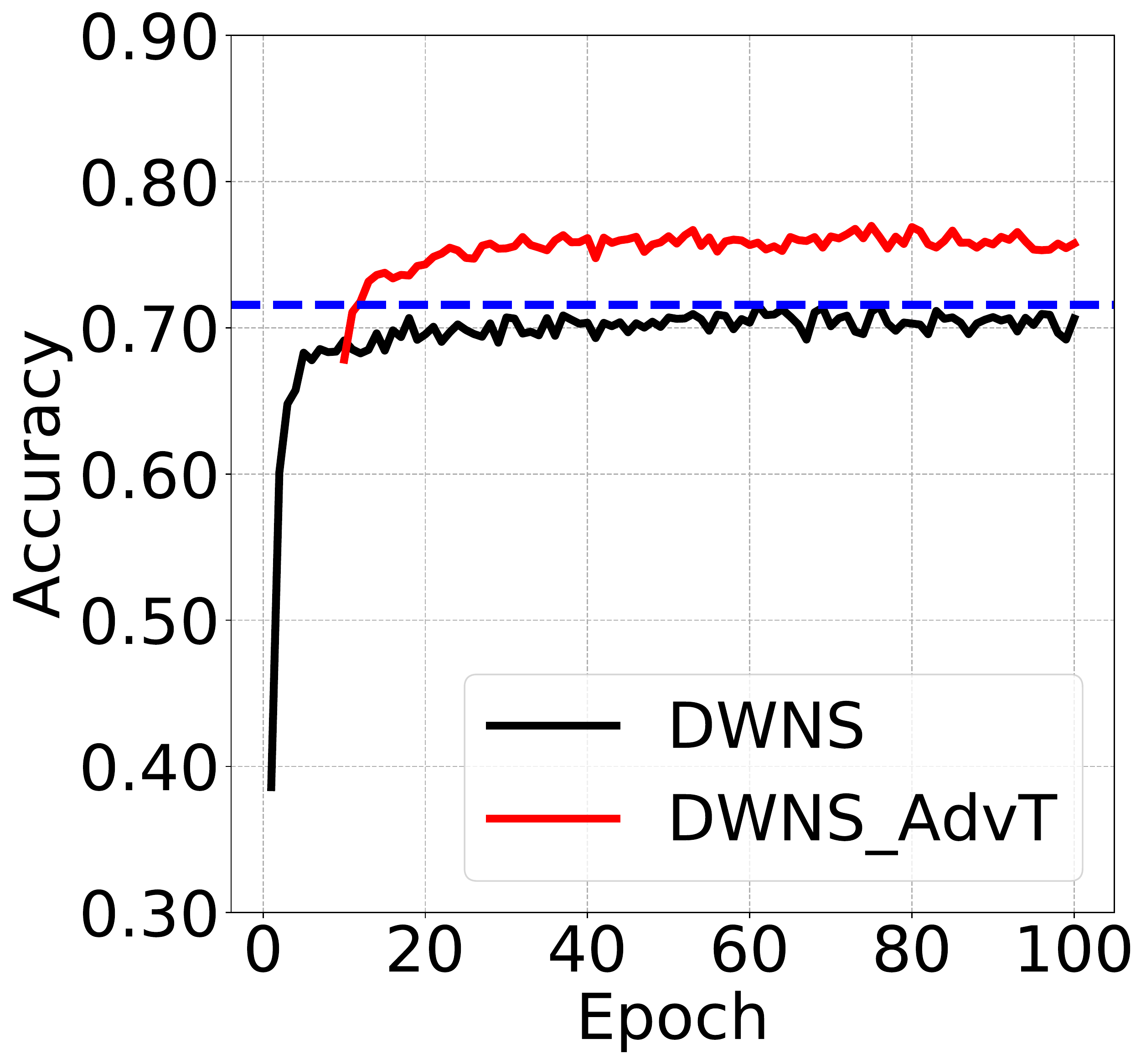}
		\hspace{-0.07in}
		\includegraphics[width=0.338\columnwidth]{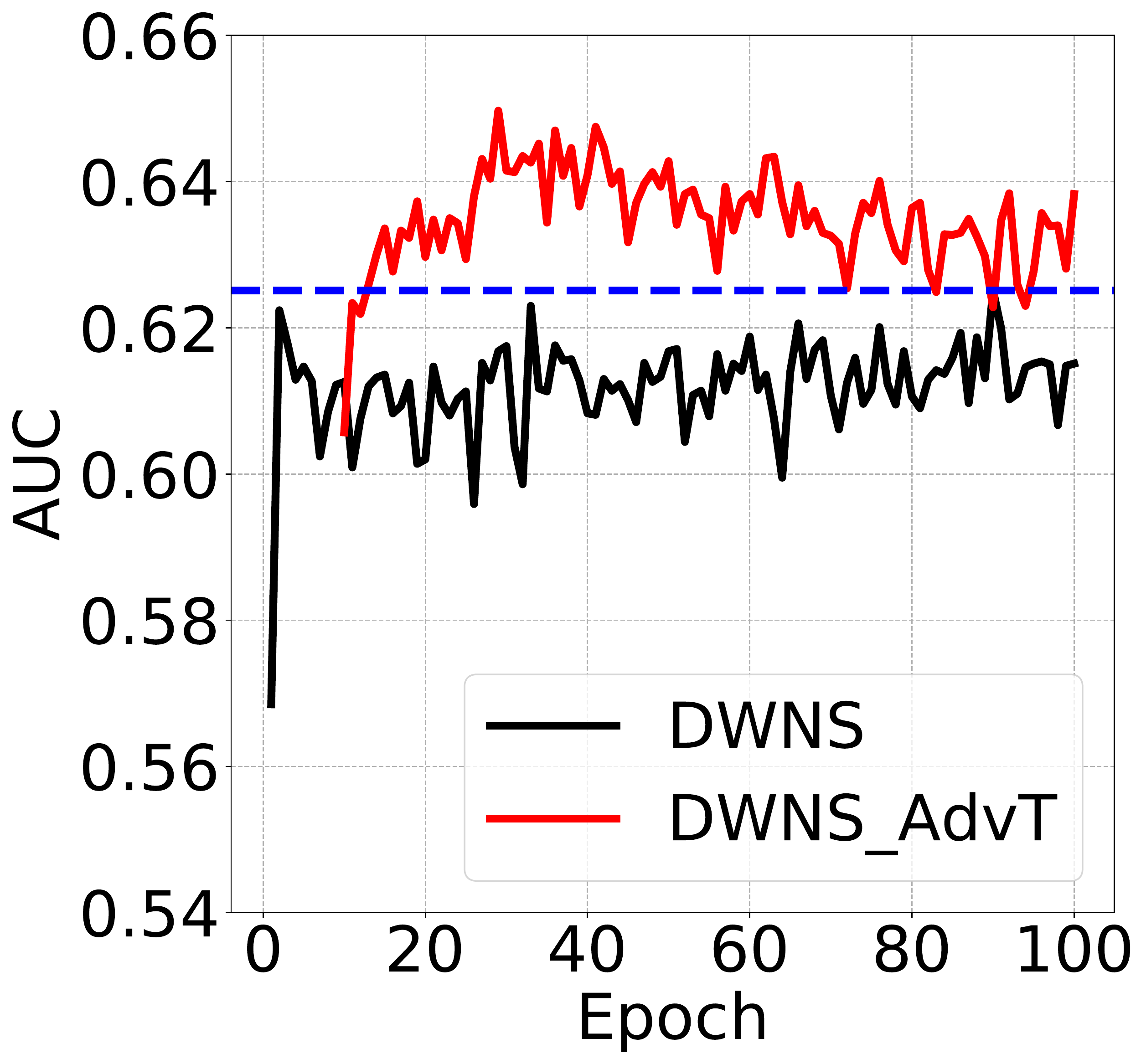}
	}
	\hspace{-0.08in}
	\subfigure[Citeseer.] { \label{fig:mcc-citeseer-adv-effect}
		\includegraphics[width=0.338\columnwidth]{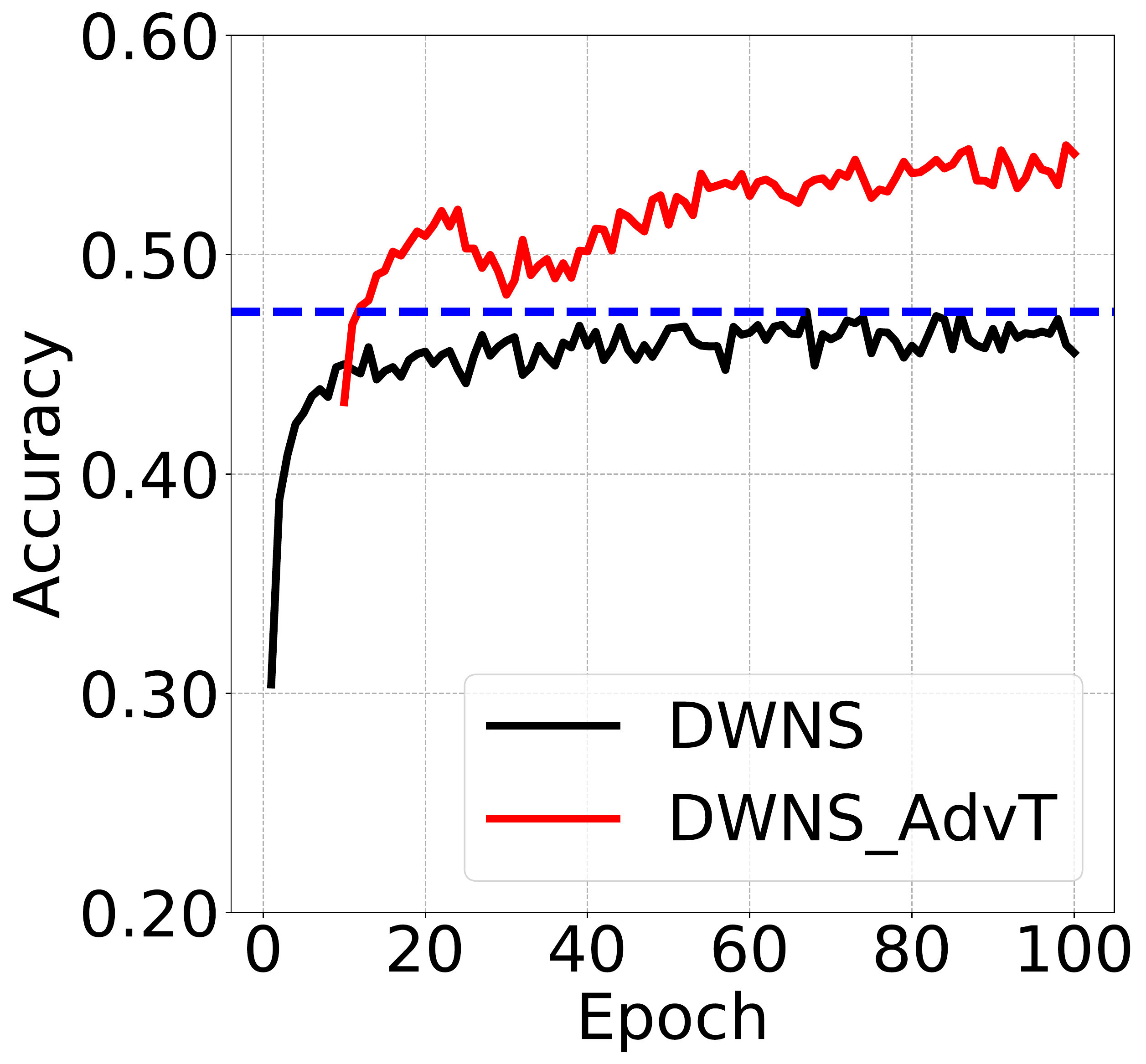}
		\hspace{-0.07in}
		\includegraphics[width=0.338\columnwidth]{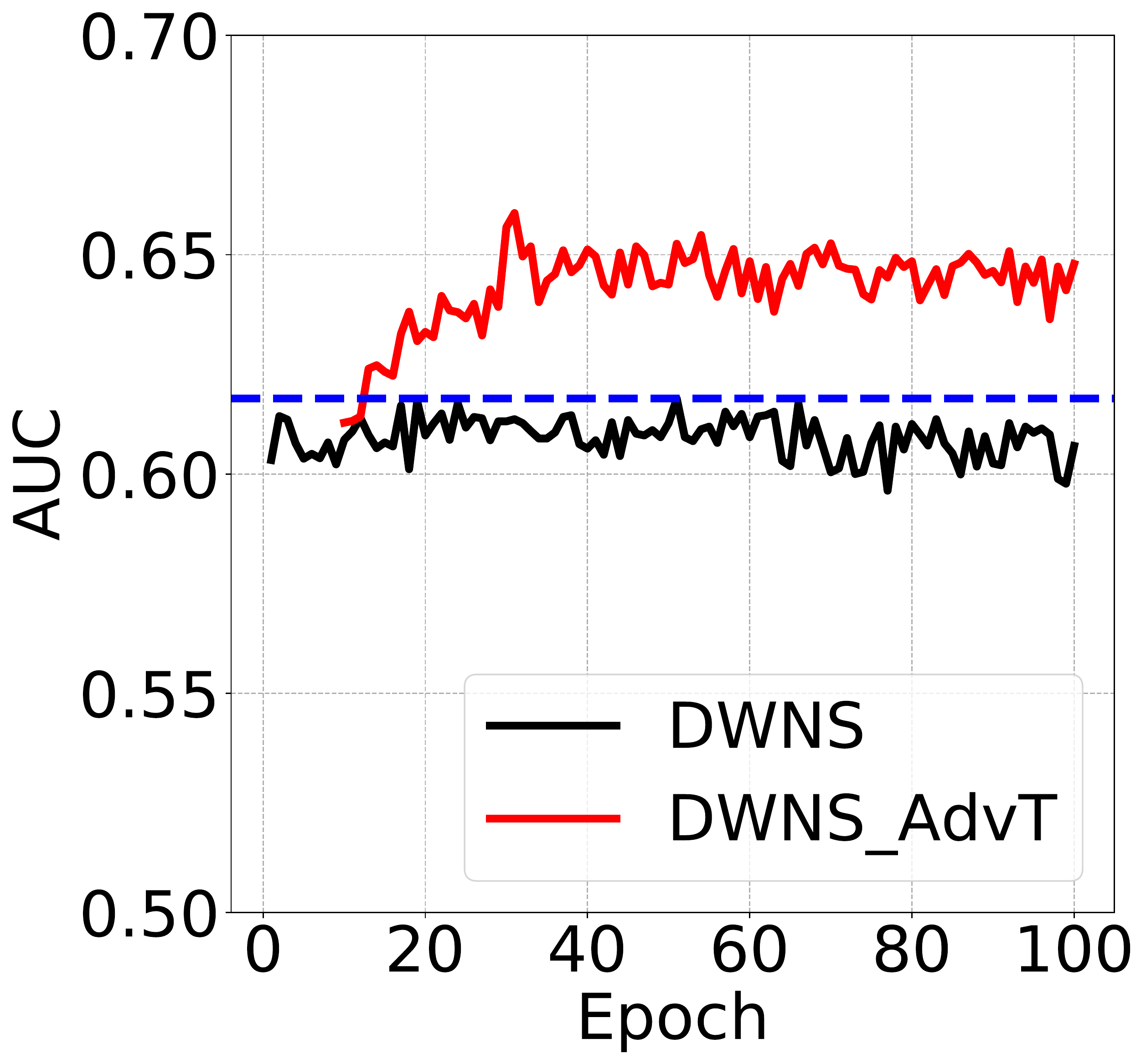}
	}
	\hspace{-0.08in}
	\subfigure[Wiki.] { \label{fig:mcc-wiki-adv-effect}
		\includegraphics[width=0.338\columnwidth]{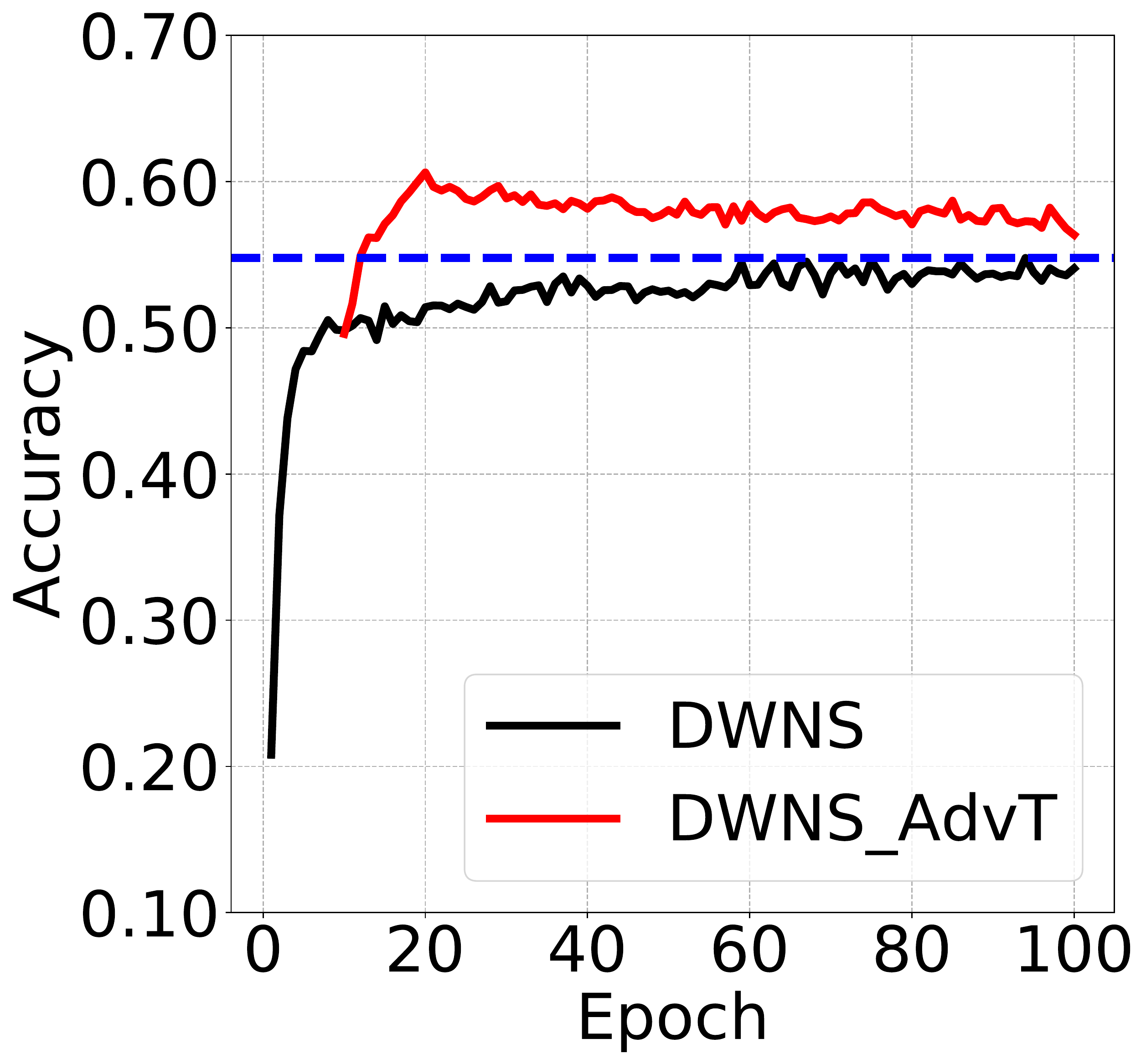}
		\hspace{-0.07in}
		\includegraphics[width=0.338\columnwidth]{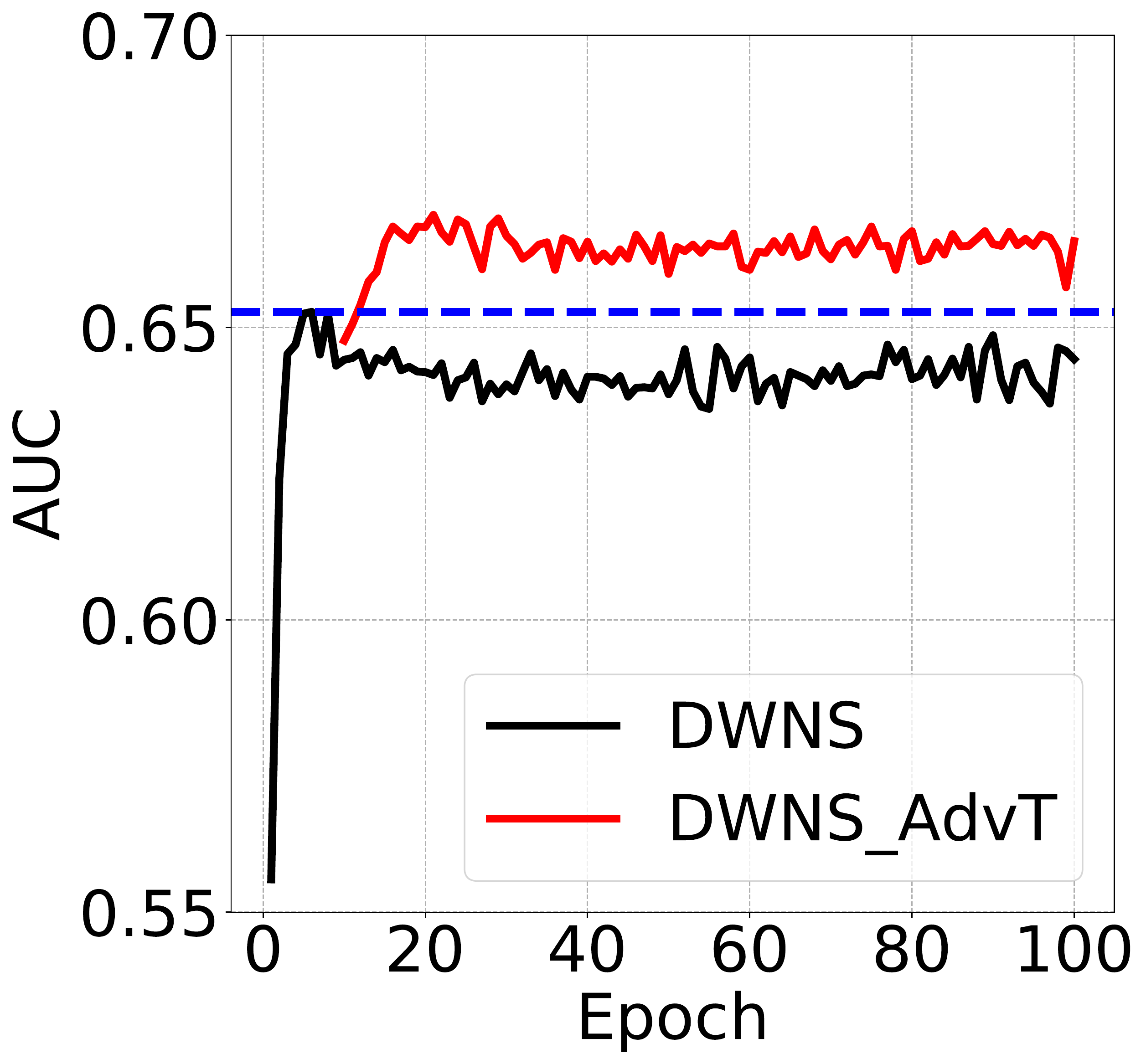}
	}
	\caption{Training curves of node classification (left, training ratio 10\%) and link prediction (right).}
	\label{fig:adv-effect}
\end{figure*}

\begin{figure*} \centering
	\subfigure[Cora, training ratio=10\%, 50\%.] { \label{fig:mcc-cora-embed-size}
		\includegraphics[width=0.338\columnwidth]{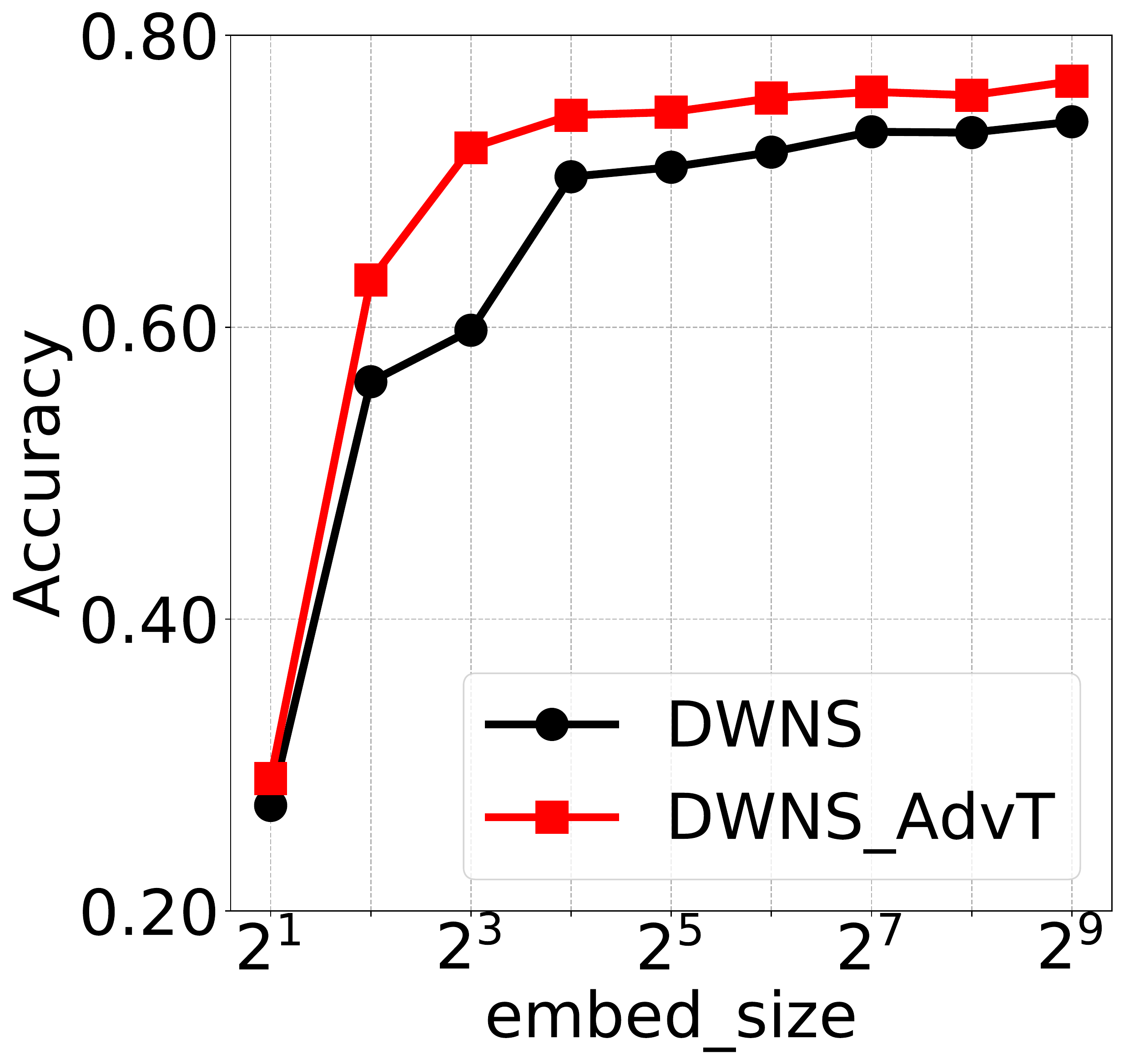}
		\hspace{-0.07in}
		\includegraphics[width=0.338\columnwidth]{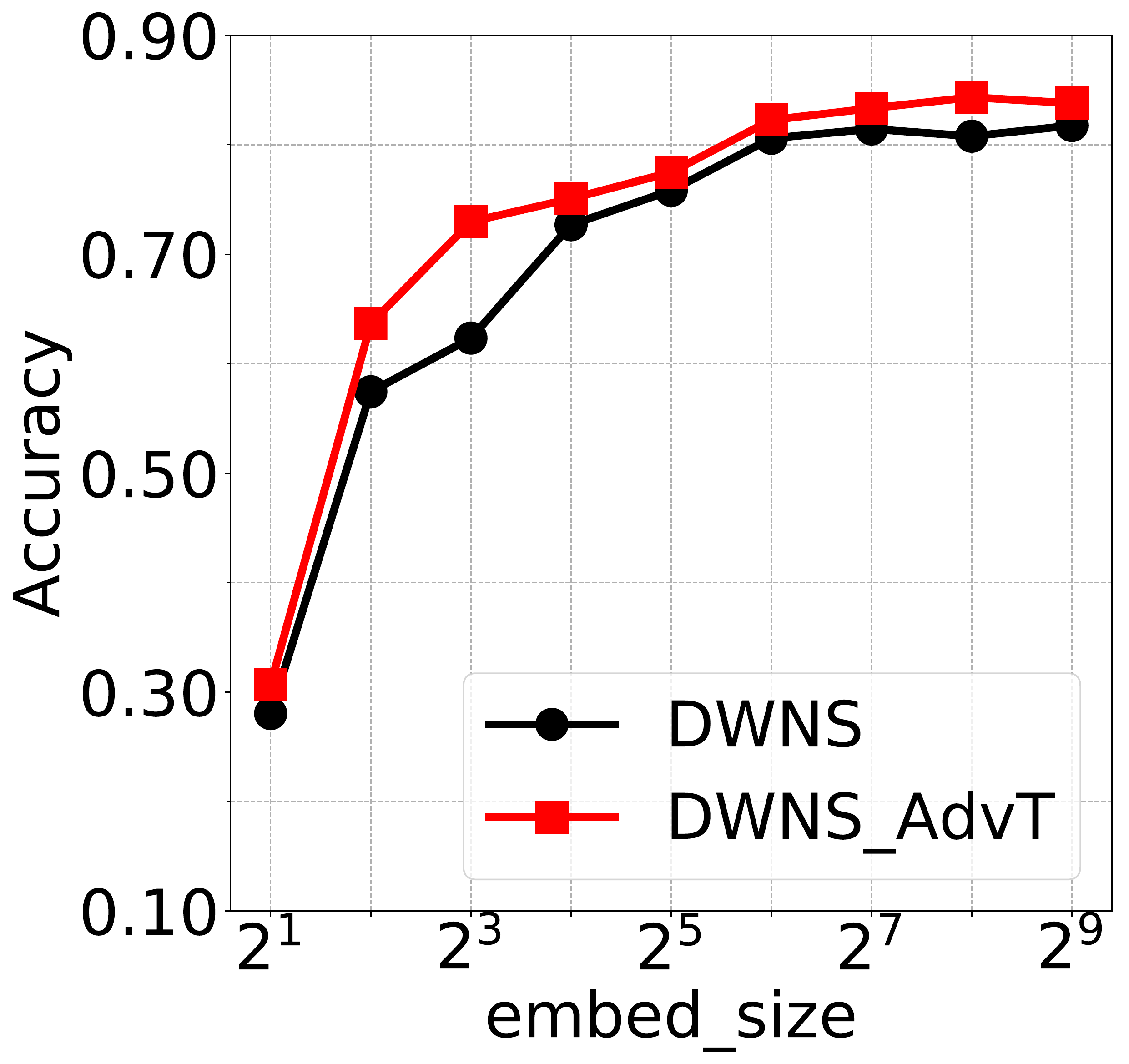}
	}
	\hspace{-0.08in}
	\subfigure[Citeseer, training ratio=10\%, 50\%.] { \label{fig:mcc-citeseer-embed-size}
		\includegraphics[width=0.338\columnwidth]{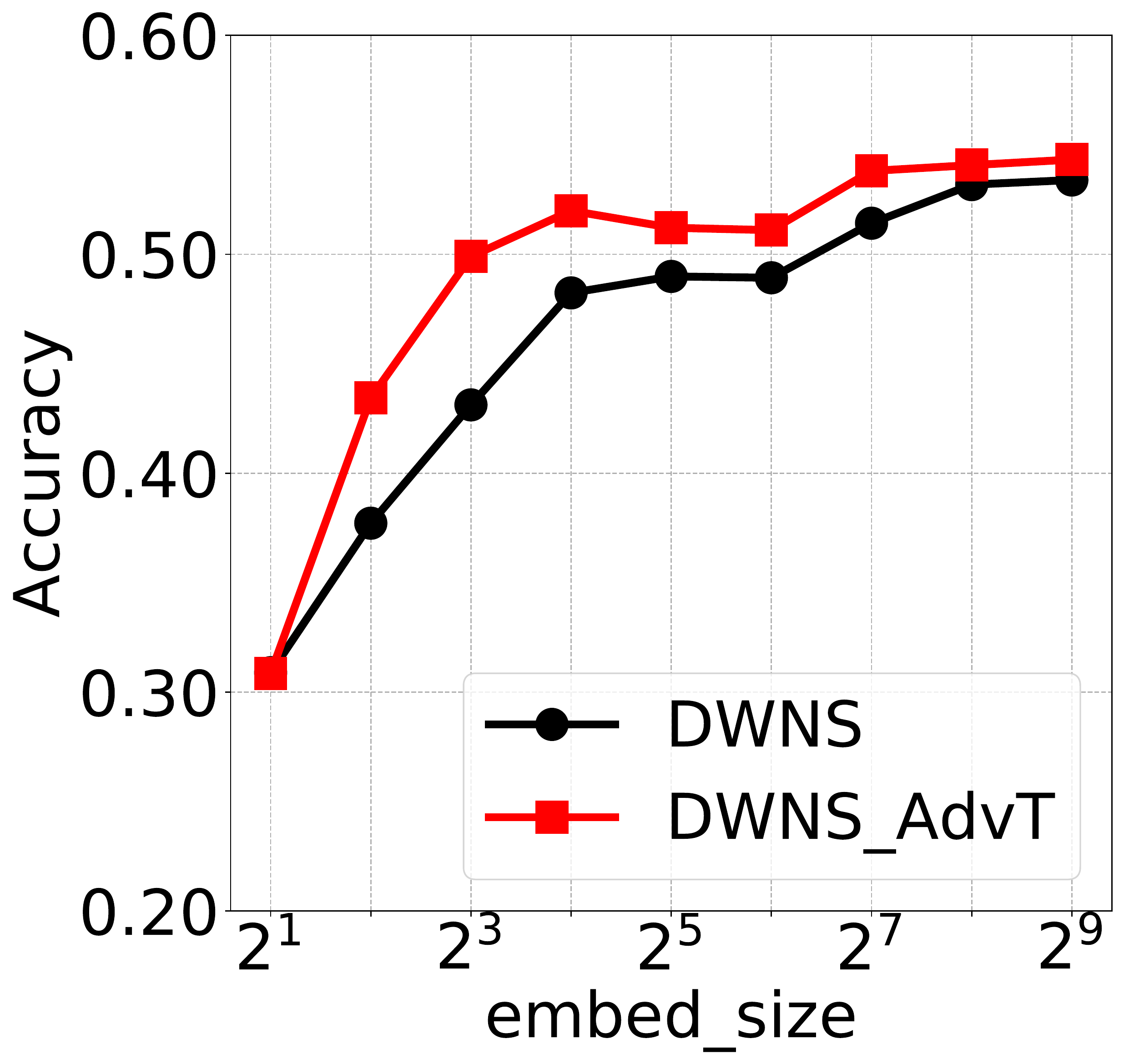}
		\hspace{-0.07in}
		\includegraphics[width=0.338\columnwidth]{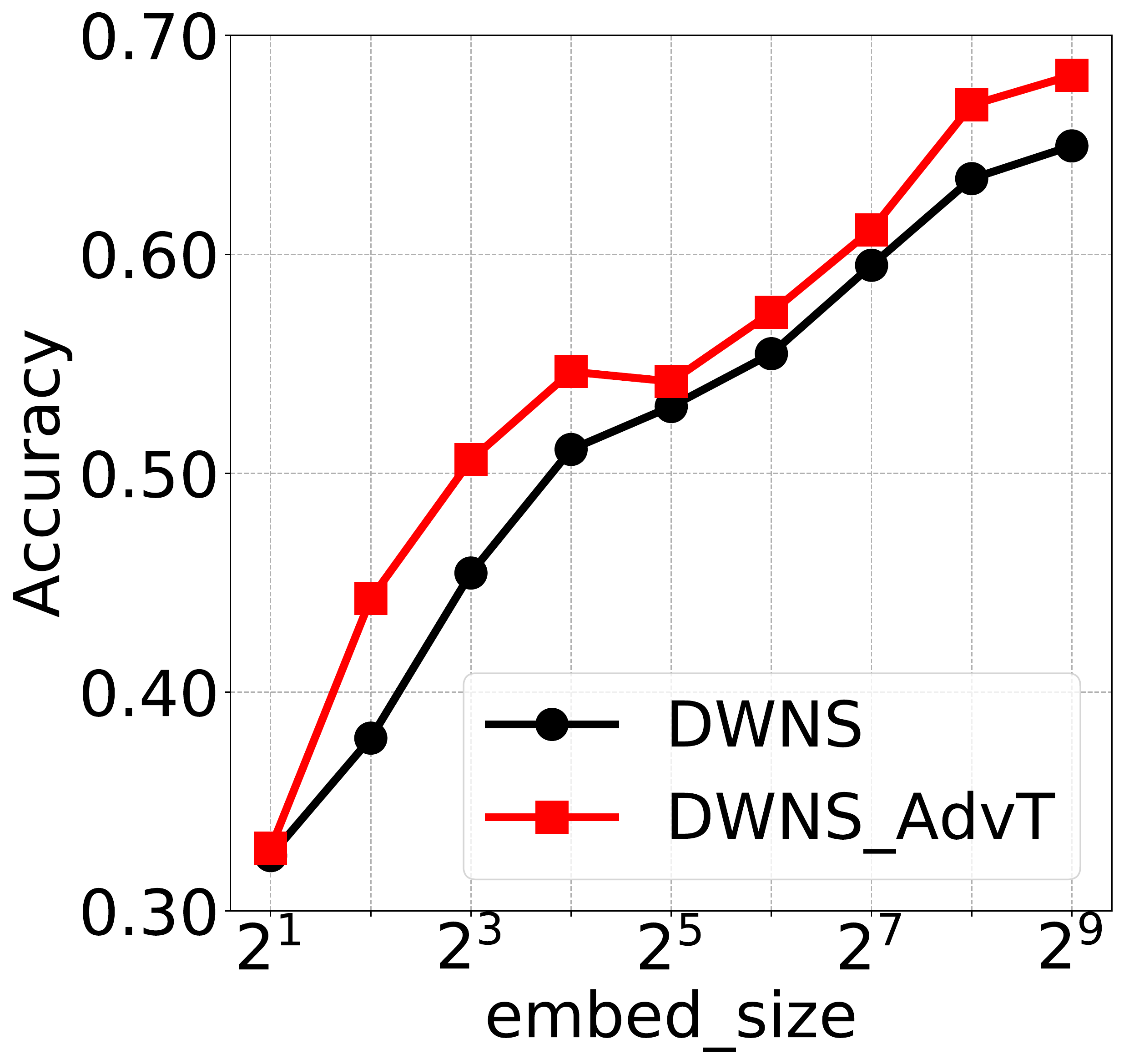}
	}
	\hspace{-0.08in}
	\subfigure[Wiki, training ratio=10\%, 50\%.] { \label{fig:mcc-wiki-embed-size}
		\includegraphics[width=0.338\columnwidth]{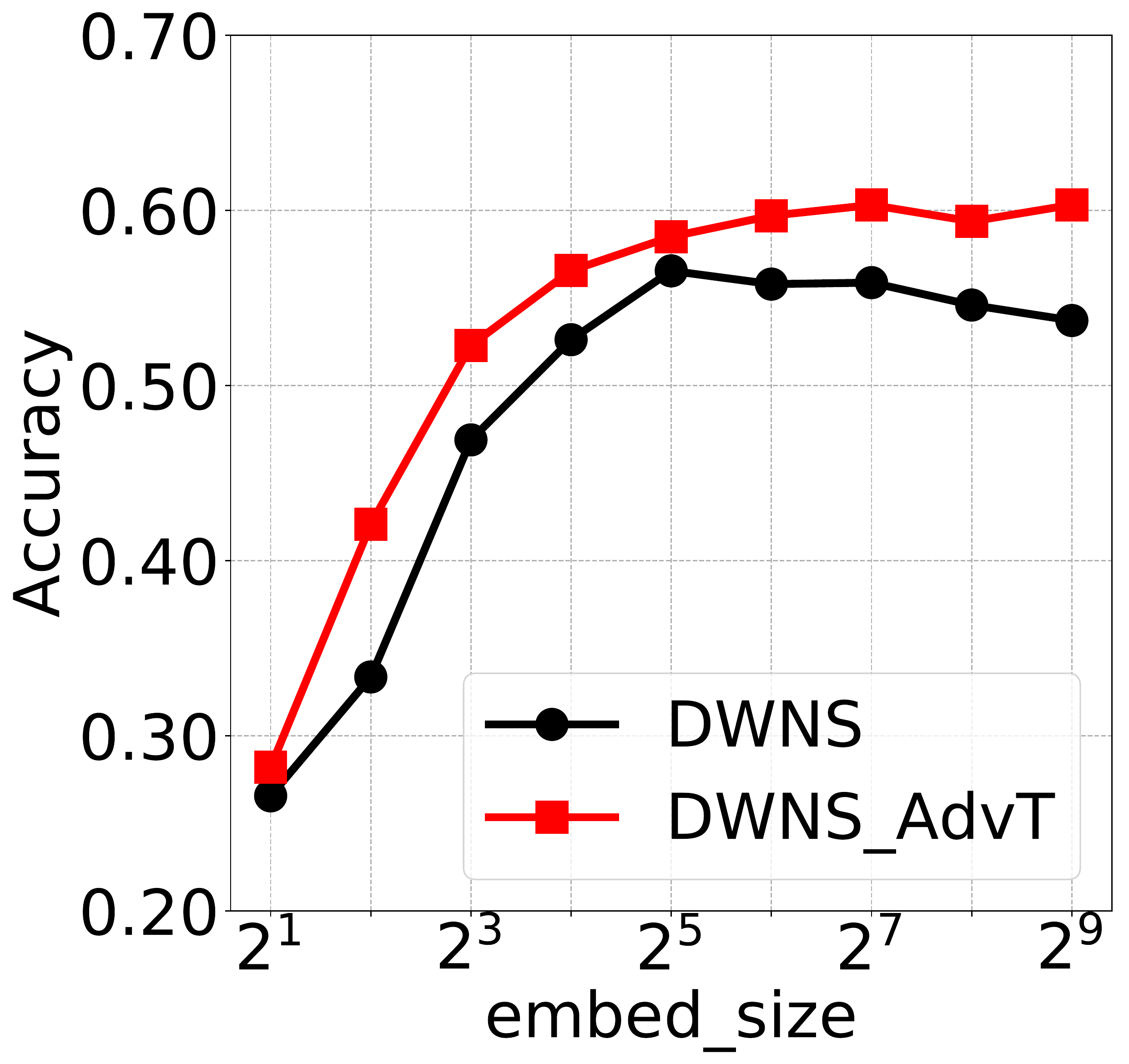}
		\hspace{-0.07in}
		\includegraphics[width=0.338\columnwidth]{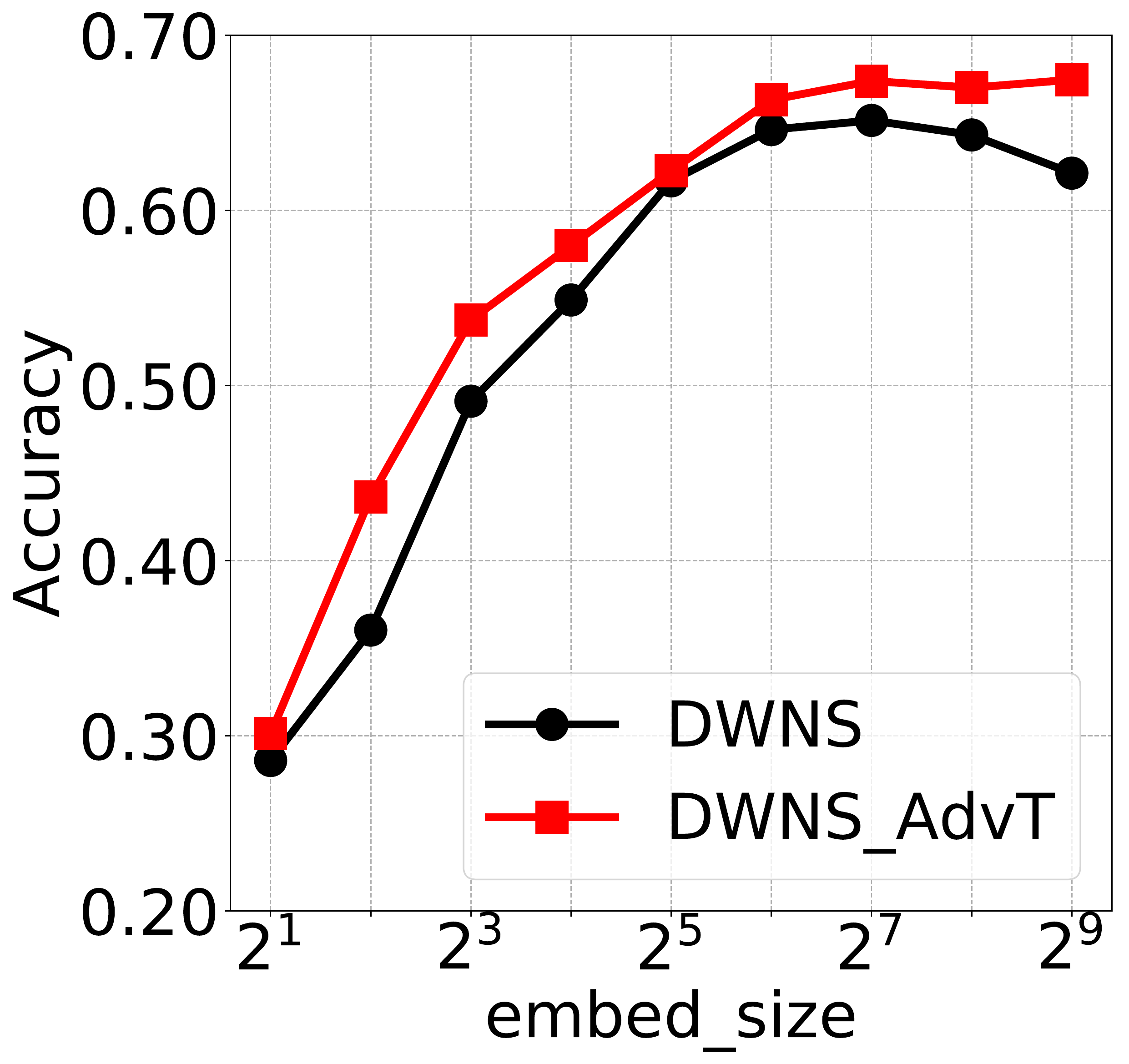}
	}
	\caption{Performance comparison between \textsc{Dwns} and \textsc{Dwns}\_AdvT on multi-class classification with training ratio as 10\% (left) and 50\% (right) respectively under varying embedding size.}
	\label{fig:embed-size}
\end{figure*}

 In this section, we conduct link prediction and multi-class classification on adversarial training \textsc{DeepWalk}, i.e., \textsc{Dwns}\_AdvT, to study the impact of adversarial training regularization on network representation learning from two aspects: model performance on different training epochs and model performance under different model size.

 Node classification is conducted with support vector classifier in Liblinear package\footnote{https://www.csie.ntu.edu.tw/$\scriptsize{\sim}$cjlin/liblinear/}~\cite{JRML-08-FanCHWL} in default settings with the learned embedding vectors as node features. 
 In link prediction, network embedding is first performed on a sub-network, which contains 80\% of edges in the original network, to learn node representations. Note that the degree of each node is ensured to be greater than or equal to 1 during subsampling process to avoid meaningless embedding vectors. We use AUC score as the performance measure, and treat link prediction as a classification problem. Specifically, a $L_2$-SVM classifier is trained with edge feature inputs obtained from the Hadamard product of embedding vectors of two endpoints as many other works~\cite{KDD-16-Grover,SIGIR-WangCL17}, positive training samples as the observed 80\% edges, and the same number of negative training samples randomly sampled from the network, i.e., node pairs without direct edge connection. The testing set consists of the hidden 20\% edges and two times of randomly sampled negative edges. All experimental results are obtained by making an average of 10 different runs.

 \subsubsection{Training Process}
 We train \textsc{Dwns} model for 100 epochs, and evaluate the generalization performance of the learned embedding vectors in each epoch with node classification and link prediction on Cora, Citeseer and Wiki. We also conduct similar experiments on \textsc{Dwns}\_AdvT for 90 epochs with the model parameters initialized from those of \textsc{Dwns} after 10$th$ training epochs. Figures~\ref{fig:adv-effect} shows the experimental results.

 In general, adversarial training regularization can bring a significant improvement in generalization ability to \textsc{Dwns} through the observation of training curves in both node classification and link prediction. Specifically, after 10 training epochs, the evaluation performance has little improvements for all datasets in two learning tasks with further training for \textsc{Dwns}, while adversarial training regularization leads to an obvious performance increase. In Figure~\ref{fig:adv-effect}, the blue line is drew by setting its vertical coordinates as the maximum value of the metrics of \textsc{Dwns} in the corresponding experiments. We can find that the training curve of \textsc{Dwns}\_AdvT is continuously above the blue line in different training epochs. Particularly, there is an impressive 7.2\% and 9.2\% relative performance improvement in link prediction for Cora and Citeseer respectively. We notice that the performance of \textsc{Dwns}\_AdvT drops slightly after about 40 training epochs for Cora in link prediction, and about 20 training epochs for Wiki in node classification. The reason might be that some networks are more vulnerable to overfitting, and deeper understanding of this phenomenon needs further exploration.

 \subsubsection{Performance vs. Embedding Size}

 We explore the effect of adversarial regularization under different model size with multi-class classification. 
 Figure~\ref{fig:embed-size} demonstrates the classification results on Cora, Citeseer and Wiki with training ratio as 10\% and 50\%. In general, adversarial training regularization is essential for improving model generalization ability. Across all tested embedding size, our proposed adversarial training \textsc{DeepWalk} can consistently outperform the base model. For two models, when varying embedding size from $2$ to $512$, the classification accuracy firstly increases in a relatively fast speed, then grows slowly, and finally becomes stable or even drops slightly. The reason is that model generalization ability is improved with the increase of model capacity firstly until some threshold, since more network structural information can be captured with larger model capacity. However, when the model capacity becomes too large, it can easily result in overfitting, and thus cause performance degradation. We notice that the performance improvement of \textsc{Dwns}\_AdvT over \textsc{Dwns} is quite small when the embedding size is 2. It is probably because model capacity is the main reason limiting model performance and model robustness is not a serious issue when embedding size is too small.

\subsection{Link Prediction}

 Link prediction is essential for many applications such as extracting missing information and identifying spurious interaction~\cite{Lv20111150}. In this section, we conduct link prediction on five real-world networks, and compare our proposed methods with the state-of-the-art methods. The experimental settings have been illustrated in Section~\ref{sec-adv-effect}. Table~\ref{tab:linkPred} summarizes the experimental results.
 
 It can be easily observed that both our proposed methods, including \textsc{Dwns}\_AdvT and \textsc{Dwns}\_iAdvT, performs better than \textsc{Dwns} in all five datasets, which demonstrates that two types of adversarial regularization methods can help improve model generalization ability. Specifically, there is a 4.62\% performance improvement for \textsc{Dwns}\_AdvT over \textsc{Dwns} on average across all datasets, and that for \textsc{Dwns}\_iAdvT is 4.60\%, which are very impressive. 
 
 We noticed that AIDW has a poor performance in link prediction. The reasons can be two-folds: firstly, AIDW encourages the smoothness of embedding distribution from a global perspective by imposing a prior distribution on them, which can result in over-regularization and thus cause performance degradation; secondly, AIDW suffers from mode-collapse problem because of its generative adversarial network component, which can also result in model corruption. Besides, \textsc{Dwns}\_rand has similar performance with \textsc{Dwns}, which means that the regularization term with random perturbation contributes little to model generalization ability. By comparison, our proposed novel adversarial training regularization method is more stable and effective.
 
 It can be observed that the performance of \textsc{Dwns}\_AdvT and \textsc{Dwns}\_iAdvT are comparable. Either \textsc{Dwns}\_AdvT or \textsc{Dwns}\_iAdvT achieves the best results across the five datasets, which shows the remarkable usefulness of the proposed regularization methods. For Cora and CA-GrQc, \textsc{Dwns}\_iAdvT has better performance, although we restrict the perturbation directions toward the nearest neighbors of the considered node. It suggests that such restriction of perturbation directions might provide useful information for representation learning. 

\subsection{Node Classification}
 Node classification can be conducted to dig out missing information in a network. In this section, we conduct multi-class classification on three benchmark datasets, including Cora, Citeseer and Wiki, with the training ratio ranging from 1\% to 90\%. Tables~\ref{tab-cora},~\ref{tab-citeseer} and~\ref{tab-wiki} summarize the experimental results.

 Firstly, \textsc{Dwns}\_rand and \textsc{Dwns} have similar performance in all three datasets. For example, the average improvement of \textsc{Dwns}\_rand over \textsc{Dwns} is 0.16\% across all training ratios in Wiki, which can be negligible. It validates that random perturbation for the regularization term contributes little to the model generalization performance again. It is understandable, since the expected dot product between any reference vector and the random perturbation from a zero mean gaussian distribution is zero, and thus the regularization term will barely affect the embedding learning. 
 
 Secondly, \textsc{Dwns}\_AdvT and \textsc{Dwns}\_iAdvT consistently outperform \textsc{Dwns} across all different training ratios in the three datasets, with the only exception of \textsc{Dwns}\_iAdvT in Citeseer when the training ratio is 3\%. Specifically, \textsc{Dwns}\_AdvT achieves 5.06\%, 6.45\% and 5.21\% performance gain over \textsc{Dwns} on average across all training ratios in Cora, Citeseer and Wiki respectively, while the improvement over \textsc{Dwns} for \textsc{Dwns}\_iAdvT are 2.35\%, 4.50\% and 2.62\% respectively. It validates that adversarial perturbation can provide useful direction for generating adversarial examples, and thus brings significant improvements to model generalization ability after the adversarial training process. For \textsc{Dwns}\_iAdvT, it brings less performance gain compared with \textsc{Dwns}\_AdvT, which might because the restriction on perturbation direction limit its regularization ability for classification tasks. In this case, there is a tradeoff between interpretability and regularization effect.
 
 Thirdly, AIDW achieves better results than \textsc{DeepWalk}, LINE and GraRep, which shows that global regularization on embedding vectors through adversarial learning can help improve model generalization performance. Our proposed methods, especially \textsc{Dwns}\_AdvT, demonstrate superiority over all the state-of-the-art baselines, including AIDW and node2vec, based on experimental results comparison. We can summarize that the adversarial training regularization method has advantages over the GAN-based global regularization methods in three aspects, including more succinct architecture, better computational efficiency and more effective performance contribution.

\begin{table*}[t]
	\centering
	\caption{AUC score for link prediction}
	\scalebox{0.85}{
		\begin{tabular}{ c | c | c | c | c | c }
			\hline
			\hline
			Dataset & Cora & Citeseer & Wiki & CA-GrQc & CA-HepTh \\
			\hline
			GF & 0.550 $\pm$ 0.005 & 0.550 $\pm$ 0.002 & 0.584 $\pm$ 0.007 & 0.593 $\pm$ 0.003 & 0.554 $\pm$ 0.001 \\
			\textsc{DeepWalk} & 0.620 $\pm$ 0.003 & 0.621 $\pm$ 0.002 & 0.658 $\pm$ 0.002 & 0.694 $\pm$ 0.001 & 0.683 $\pm$ 0.000 \\
			LINE & 0.626 $\pm$ 0.011 & 0.625 $\pm$ 0.004 & 0.647 $\pm$ 0.010 & 0.641 $\pm$ 0.002 & 0.629 $\pm$ 0.005 \\
			node2vec & 0.626 $\pm$ 0.023 & 0.627 $\pm$ 0.022 & 0.639 $\pm$ 0.010 & 0.695 $\pm$ 0.006 & 0.667 $\pm$ 0.009 \\
			GraRep & 0.609 $\pm$ 0.035 & 0.589 $\pm$ 0.025 & 0.642 $\pm$ 0.045 & 0.500 $\pm$ 0.000 & 0.500 $\pm$ 0.000 \\
			AIDW & 0.552 $\pm$ 0.034 & 0.606 $\pm$ 0.035 & 0.511 $\pm$ 0.019 & 0.615 $\pm$ 0.023 & 0.592 $\pm$ 0.019 \\
			\hline
			\textsc{Dwns} & 0.609 $\pm$ 0.018 & 0.609 $\pm$ 0.011 & 0.648 $\pm$ 0.007 & 0.690 $\pm$ 0.004 & 0.662 $\pm$ 0.006 \\
			\textsc{Dwns}\_rand & 0.606 $\pm$ 0.012 & 0.608 $\pm$ 0.005 & 0.645 $\pm$ 0.010 & 0.696 $\pm$ 0.006 & 0.662 $\pm$ 0.003\\
			\textsc{Dwns}\_AdvT & 0.644 $\pm$ 0.009 & \textbf{0.656} $\pm$ \textbf{0.007} & \textbf{0.665} $\pm$ \textbf{0.005} & \textbf{0.707} $\pm$ \textbf{0.004} & \textbf{0.692} $\pm$ \textbf{0.003} \\
			\textsc{Dwns}\_iAdvT & \textbf{0.655} $\pm$ \textbf{0.015} & 0.653 $\pm$ 0.006 & 0.660 $\pm$ 0.002 & \textbf{0.707} $\pm$ \textbf{0.004} & 0.688 $\pm$ 0.004 \\
			\hline
			\hline
	\end{tabular}}
	\label{tab:linkPred}
\end{table*}


 \begin{table*}[htb!]
 \centering
 \caption{Accuracy (\%) of multi-class classification on Cora}
 \scalebox{0.75}{
 \begin{tabular}{ c | c | c | c | c | c | c | c | c | c | c | c | c | c | c | c | c | c | c }
  \hline
  \hline
  \%Ratio & 1\% & 2\% & 3\% & 4\% & 5\% & 6\% & 7\% & 8\% & 9\% & 10\% & 20\% & 30\% & 40\% & 50\% & 60\% & 70\% & 80\% & 90\% \\
  \hline
    GF & 24.55 & 28.87 & 32.07 & 33.11 & 34.45 & 35.83 & 38.25 & 39.05 & 39.84 & 39.42 & 46.14 & 48.57 & 50.09 & 50.85 & 51.88 & 52.89 & 52.34 & 51.51 \\
    \textsc{DeepWalk} & 44.63 & 49.30 & 52.25 & 53.05 & 55.21 & 59.10 & 59.26 & 62.20 & 63.07 & 64.60 & 69.85 & 74.21 & 76.68 & 77.59 & 77.68 & 78.63 & 79.35 & 79.23 \\
    LINE & 38.78 & 49.62 & 54.51 & 56.49 & 58.99 & 61.30 & 63.05 & 64.19 & 66.59 & 66.06 & 70.86 & 72.25 & 73.94 & 74.03 & 74.65 & 75.12 & 75.30 & 75.76 \\
    node2vec & 58.02 & 63.98 & 66.33 & 68.07 & 69.91 & 69.87 & 71.41 & 72.60 & 73.63 & 73.96 & 78.04 & 80.07 & 81.62 & 82.16 & 82.25 & 82.85 & 84.02 & 84.91 \\
    GraRep & 54.24 & 63.58 & 65.36 & 68.78 & 70.67 & 72.69 & 72.37 & 72.70 & 73.53 & 74.98 & 77.48 & 78.57 & 79.38 & 79.53 & 79.68 & 79.75 & 80.89 & 80.74 \\
    AIDW & 54.55 & 63.30 & 65.86 & 66.20 & 67.62 & 68.61 & 69.52 & 71.07 & 71.44 & 73.83 & 77.93 & 79.43 & 81.16 & 81.79 & 82.27 & 82.93 & 84.11 & 83.69 \\
  \hline
    \textsc{Dwns} & 57.72 & 64.82 & 67.93 & 68.50 & 68.27 & 70.81 & 70.72 & 72.30 & 72.00 & 73.20 & 76.98 & 79.83 & 80.56 & 82.27 & 82.52 & 82.92 & 82.97 & 84.54 \\
    \textsc{Dwns}\_rand & 56.46 & 64.87 & 67.44 & 68.24 & 70.38 & 71.16 & 71.34 & 72.67 & 73.51 & 73.45 & 78.04 & 79.76 & 81.66 & 81.72 & 82.53 & 83.57 & 83.51 & 83.69 \\
    \textsc{Dwns}\_AdvT & \textbf{62.66} & \textbf{68.46} & 69.91 & \textbf{73.62} & \textbf{74.71} & \textbf{75.55} & \textbf{76.18} & \textbf{76.77} & \textbf{77.72} & \textbf{77.73} & \textbf{80.50} & \textbf{82.33} & \textbf{83.54} & \textbf{83.63} & \textbf{84.41} & \textbf{84.99} & \textbf{85.66} & \textbf{85.65} \\
    \textsc{Dwns}\_iAdvT & 58.67 & 66.65 & \textbf{70.17} & 70.52 & 71.42 & 72.47 & 74.26 & 75.32 & 74.52 & 76.12 & 78.88 & 80.31 & 81.61 & 82.80 & 83.03 & 83.63 & 83.75 & 85.02 \\
  \hline
  \hline
 \end{tabular}}
 \label{tab-cora}
 \end{table*}


 \begin{table*}[htb!]
 \centering
 \caption{Accuracy (\%) of multi-class classification on Citeseer}
 \scalebox{0.75}{
 \begin{tabular}{ c | c | c | c | c | c | c | c | c | c | c | c | c | c | c | c | c | c | c }
  \hline
  \hline
  \%Ratio & 1\% & 2\% & 3\% & 4\% & 5\% & 6\% & 7\% & 8\% & 9\% & 10\% & 20\% & 30\% & 40\% & 50\% & 60\% & 70\% & 80\% & 90\% \\
  \hline
    GF & 22.63 & 24.49 & 25.76 & 28.21 & 28.07 & 29.02 & 30.20 & 30.70 & 31.20 & 31.48 & 34.05 & 35.69 & 36.26 & 37.18 & 37.87 & 38.85 & 39.16 & 39.54 \\
    \textsc{DeepWalk} & 27.82 & 32.44 & 35.47 & 36.85 & 39.10 & 41.01 & 41.56 & 42.81 & 45.35 & 45.53 & 50.98 & 53.79 & 55.25 & 56.05 & 56.84 & 57.36 & 58.15 & 59.11 \\
    LINE & 29.98 & 34.91 & 37.02 & 40.51 & 41.63 & 42.48 & 43.65 & 44.25 & 45.65 & 47.03 & 50.09 & 52.71 & 53.52 & 54.20 & 55.42 & 55.87 & 55.93 & 57.22 \\
    node2vec & 36.56 & 40.21 & 44.14 & 45.71 & 46.32 & 47.47 & 49.56 & 49.78 & 50.73 & 50.78 & 55.89 & 57.93 & 58.60 & 59.44 & 59.97 & 60.32 & 60.75 & 61.04 \\
    GraRep & 37.98 & 40.72 & 43.33 & 45.56 & 47.48 & 47.93 & 49.54 & 49.87 & 50.65 & 50.60 & 53.56 & 54.63 & 55.44 & 55.20 & 55.07 & 56.04 & 55.48 & 56.39 \\
    AIDW & 38.77 & 42.84 & 44.04 & 44.27 & 46.29 & 47.89 & 47.73 & 49.61 & 49.55 & 50.77 & 54.82 & 56.96 & 58.04 & 59.65 & 60.03 & 60.99 & 61.18 & 62.84 \\
  \hline
    \textsc{Dwns} & 38.13 & 42.88 & 46.60 & 46.14 & 46.38 & 48.18 & 48.58 & 48.35 & 50.16 & 50.00 & 53.74 & 57.37 & 58.59 & 59.00 & 59.53 & 59.62 & 59.51 & 60.18 \\
    \textsc{Dwns}\_rand & 39.29 & 43.42 & 42.73 & 46.00 & 46.13 & 48.69 & 48.15 & 49.92 & 50.08 & 50.84 & 55.26 & 58.51 & 59.59 & 59.12 & 60.22 & 60.62 & 61.59 & 60.55 \\
    \textsc{Dwns}\_AdvT & \textbf{41.33} & 45.00 & \textbf{46.73} & \textbf{48.57} & \textbf{50.37} & \textbf{51.06} & \textbf{52.07} & \textbf{53.09} & \textbf{53.73} & \textbf{54.79} & \textbf{59.21} & \textbf{61.06} & \textbf{61.26} & \textbf{62.56} & \textbf{62.63} & \textbf{62.40} & \textbf{63.05} & \textbf{63.73} \\
    \textsc{Dwns}\_iAdvT & 40.88 & \textbf{45.53} & 46.01 & 47.10 & 50.02 & 50.79 & 49.59 & 52.78 & 51.95 & 52.26 & 56.65 & 59.07 & 60.27 & 61.96 & 62.04 & 62.20 & 62.21 & 63.15 \\
  \hline
  \hline
 \end{tabular}}
 \label{tab-citeseer}
 \end{table*}


 \begin{table*}[htb!]
 \centering
 \caption{Accuracy (\%) of multi-class classification on Wiki}
 \scalebox{0.75}{
 \begin{tabular}{ c | c | c | c | c | c | c | c | c | c | c | c | c | c | c | c | c | c | c }
  \hline
  \hline
  \%Ratio & 1\% & 2\% & 3\% & 4\% & 5\% & 6\% & 7\% & 8\% & 9\% & 10\% & 20\% & 30\% & 40\% & 50\% & 60\% & 70\% & 80\% & 90\% \\
  \hline
    GF & 19.76 & 22.70 & 27.00 & 28.41 & 30.28 & 31.49 & 31.87 & 32.18 & 34.16 & 34.25 & 36.13 & 37.66 & 37.43 & 39.48 & 40.17 & 39.83 & 40.25 & 41.01 \\
    \textsc{DeepWalk} & 28.65 & 32.84 & 36.66 & 37.98 & 40.73 & 42.94 & 45.57 & 45.47 & 46.06 & 46.60 & 54.48 & 59.05 & 62.70 & 64.66 & 65.95 & 66.98 & 68.37 & 68.78 \\
    LINE & 32.46 & 40.84 & 44.56 & 49.59 & 51.11 & 52.37 & 54.32 & 55.72 & 56.51 & 57.88 & 61.08 & 63.50 & 64.68 & 66.29 & 66.91 & 67.43 & 67.46 & 68.61 \\
    node2vec & 32.41 & 41.96 & 47.32 & 48.15 & 50.65 & 51.08 & 52.71 & 54.66 & 54.81 & 55.94 & 59.67 & 61.11 & 64.21 & 65.08 & 65.58 & 66.76 & 67.19 & 68.73 \\
    GraRep & 33.38 & 45.61 & 49.10 & 50.92 & 53.01 & 54.43 & 54.84 & 57.50 & 57.01 & 58.57 & 61.91 & 63.58 & 63.77 & 64.68 & 65.39 & 65.92 & 65.18 & 67.05 \\
    AIDW & 35.17 & 43.05 & 46.63 & 51.29 & 52.40 & 52.72 & 55.92 & 56.78 & 55.92 & 57.32 & 61.84 & 63.54 & 64.90 & 65.58 & 66.54 & 65.59 & 66.58 & 68.02 \\
  \hline
    \textsc{Dwns} & 35.76 & 42.71 & 48.08 & 50.01 & 50.21 & 52.26 & 53.26 & 53.80 & 55.27 & 55.77 & 59.63 & 61.98 & 64.01 & 64.59 & 66.99 & 66.45 & 67.55 & 67.51 \\
    \textsc{Dwns}\_rand & 36.12 & 44.57 & 46.71 & 49.15 & 51.74 & 53.37 & 53.22 & 53.27 & 54.21 & 56.33 & 59.41 & 61.94 & 64.07 & 65.17 & 66.18 & 65.64 & 68.20 & 67.34 \\
    \textsc{Dwns}\_AdvT & \textbf{38.42} & \textbf{45.80} & \textbf{50.21} & 51.12 & \textbf{54.29} & \textbf{56.43} & \textbf{57.12} & \textbf{57.82} & \textbf{58.60} & \textbf{59.97} & \textbf{63.33} & \textbf{65.32} & \textbf{66.53} & \textbf{67.06} & \textbf{67.69} & \textbf{68.94} & \textbf{68.35} & \textbf{69.32} \\
    \textsc{Dwns}\_iAdvT & 37.46 & 45.11 & 49.14 & \textbf{51.57} & 51.88 & 54.43 & 55.42 & 56.05 & 55.93 & 57.81 & 61.40 & 63.37 & 65.71 & 65.56 & 67.09 & 66.81 & 67.70 & 68.02 \\
  \hline
  \hline
 \end{tabular}}
 \label{tab-wiki}
 \end{table*}

 \subsection{Parameter Sensitivity}

 We conduct parameter sensitivity analysis with link prediction and multi-class classification on Cora, Citeseer and Wiki in this section. 
 Here we only present the results for \textsc{Dwns}\_AdvT due to space limitation. Adversarial training regularization method is very succinct. \textsc{Dwns}\_AdvT only has two more hyperparameters compared with \textsc{Dwns}, which are noise level $\epsilon$ and adversarial regularization strength $\lambda$. Note that when studying one hyper-parameter, we follow default settings for other hyper-parameters. The experimental settings of link prediction and node classification have been explained in Section~\ref{sec-adv-effect}.
 
\begin{figure} \centering
	\subfigure[Noise level $\epsilon$.] { \label{fig:eps}
		\includegraphics[width=0.46\columnwidth]{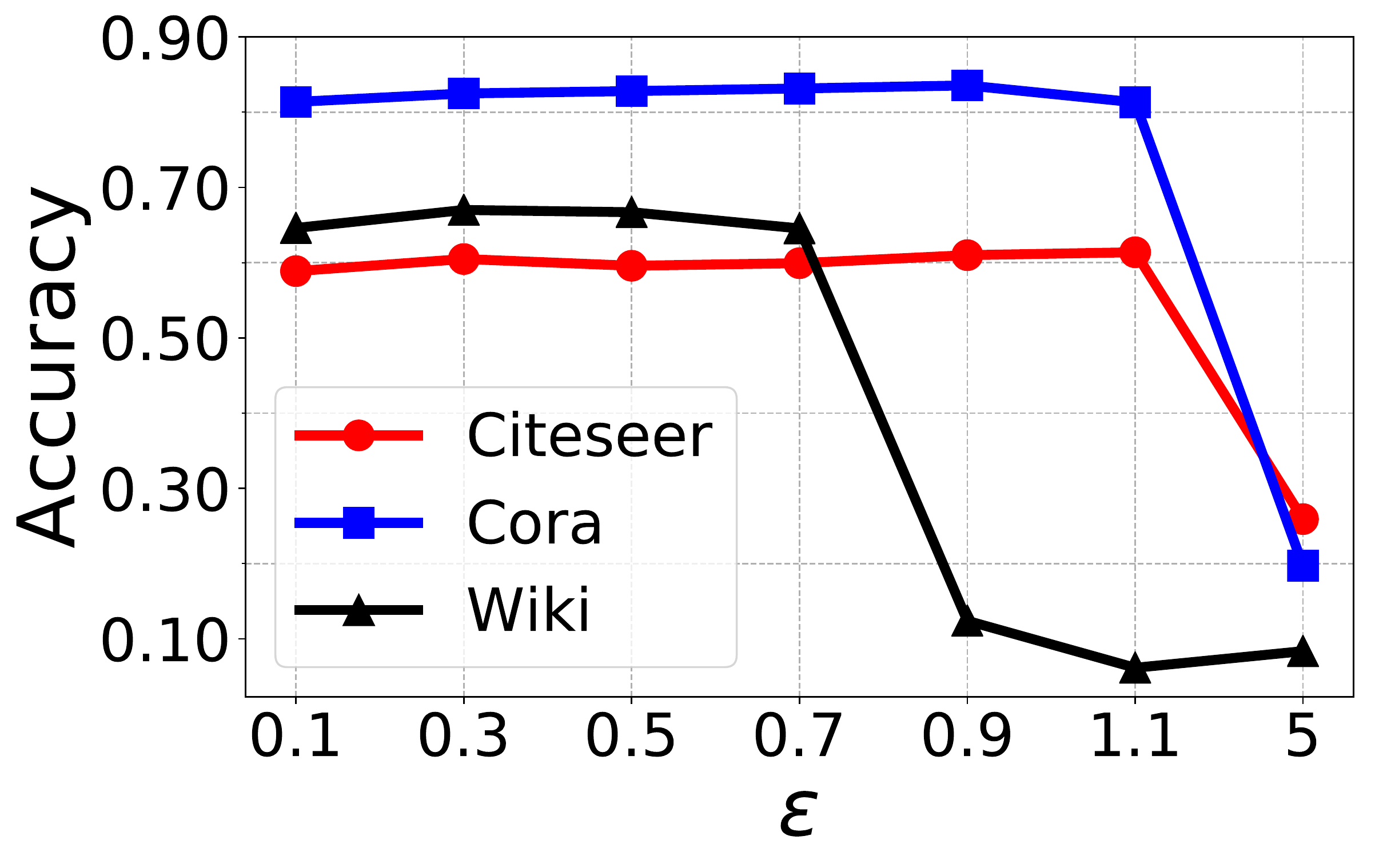}
		\hspace{0in}
		\includegraphics[width=0.46\columnwidth]{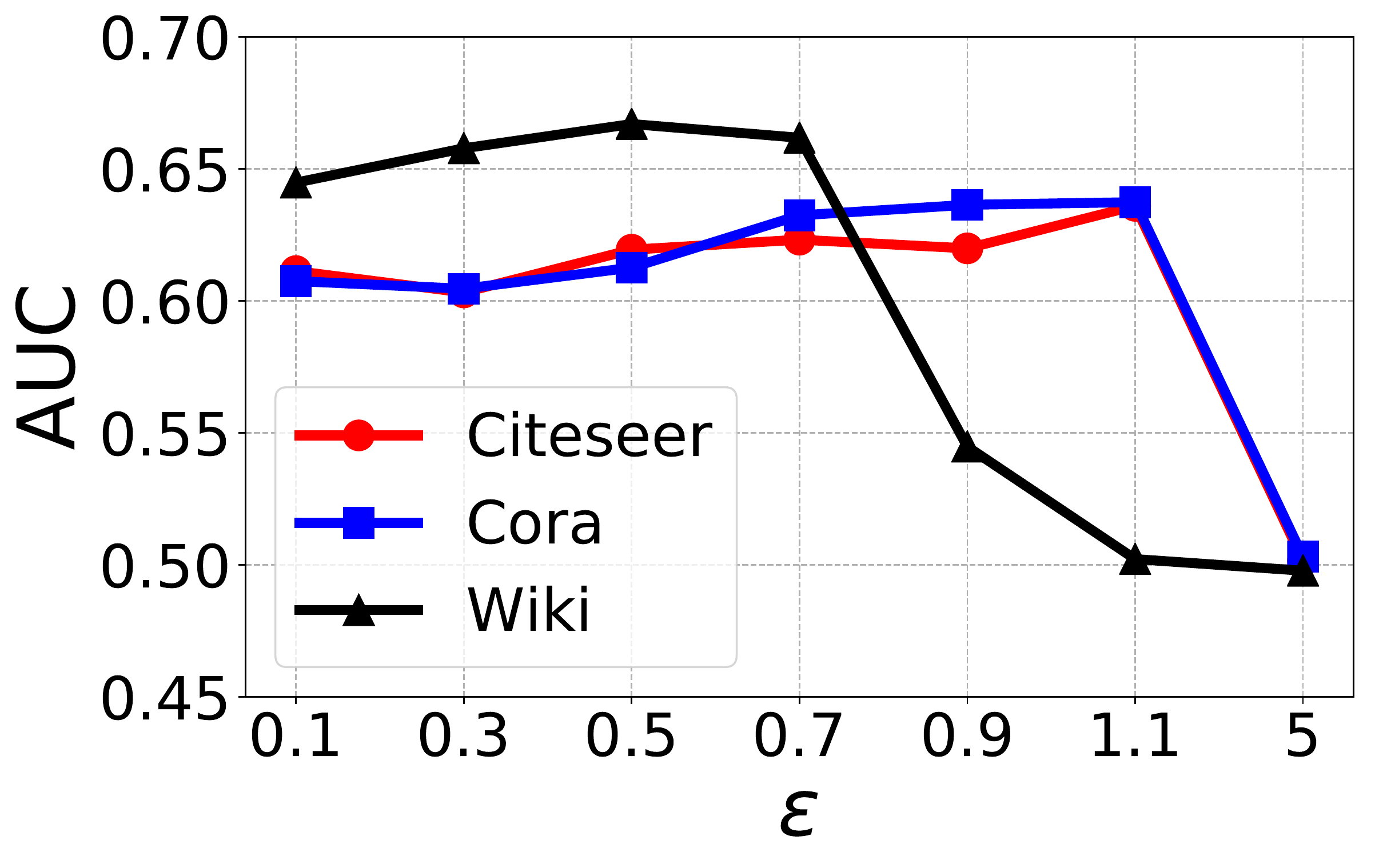}
	}
	\hspace{0.1in}
	\subfigure[Adversarial regularization strength $\lambda$.] { \label{fig:reg_adv}
		\includegraphics[width=0.46\columnwidth]{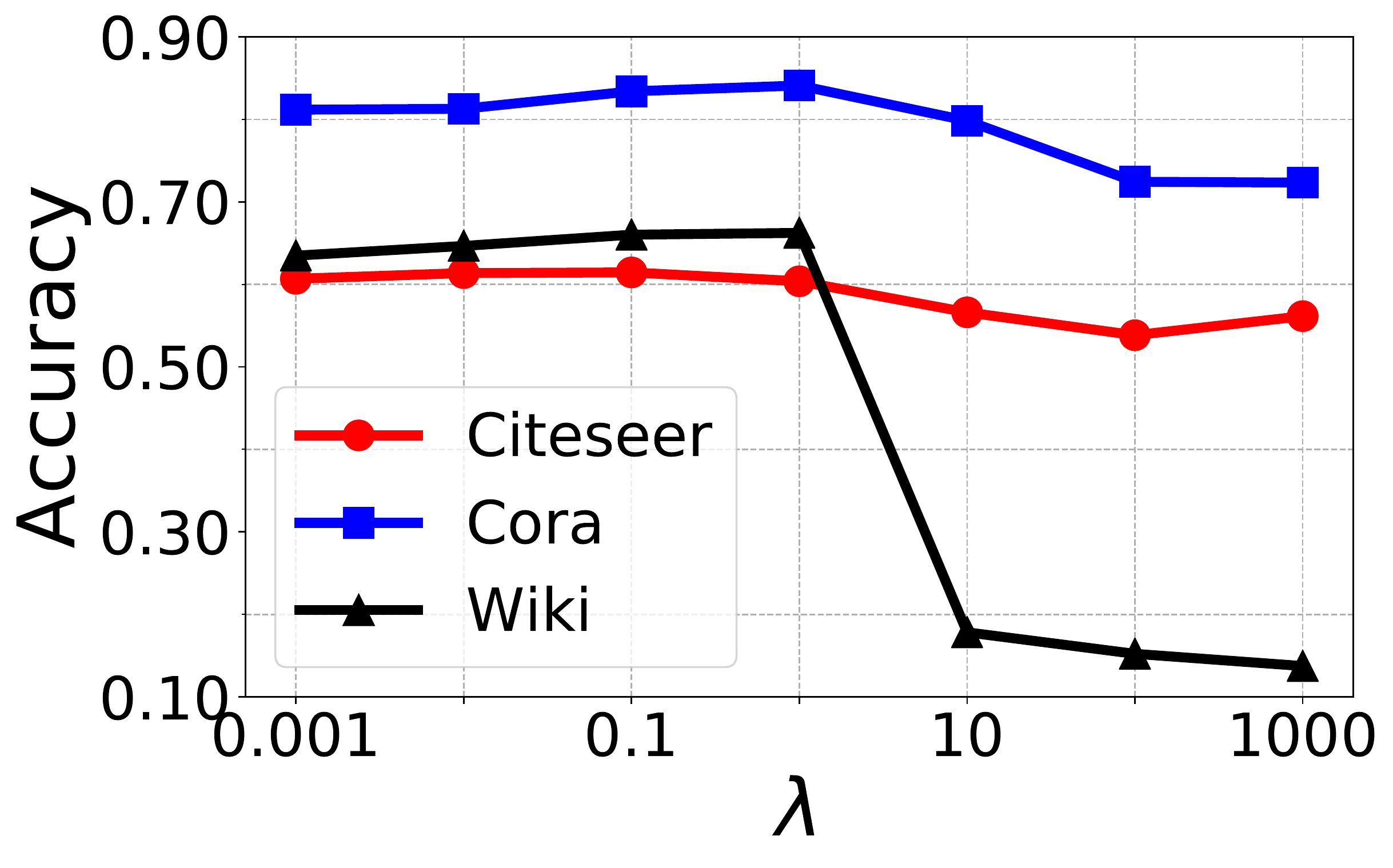}
		\hspace{0in}
		\includegraphics[width=0.46\columnwidth]{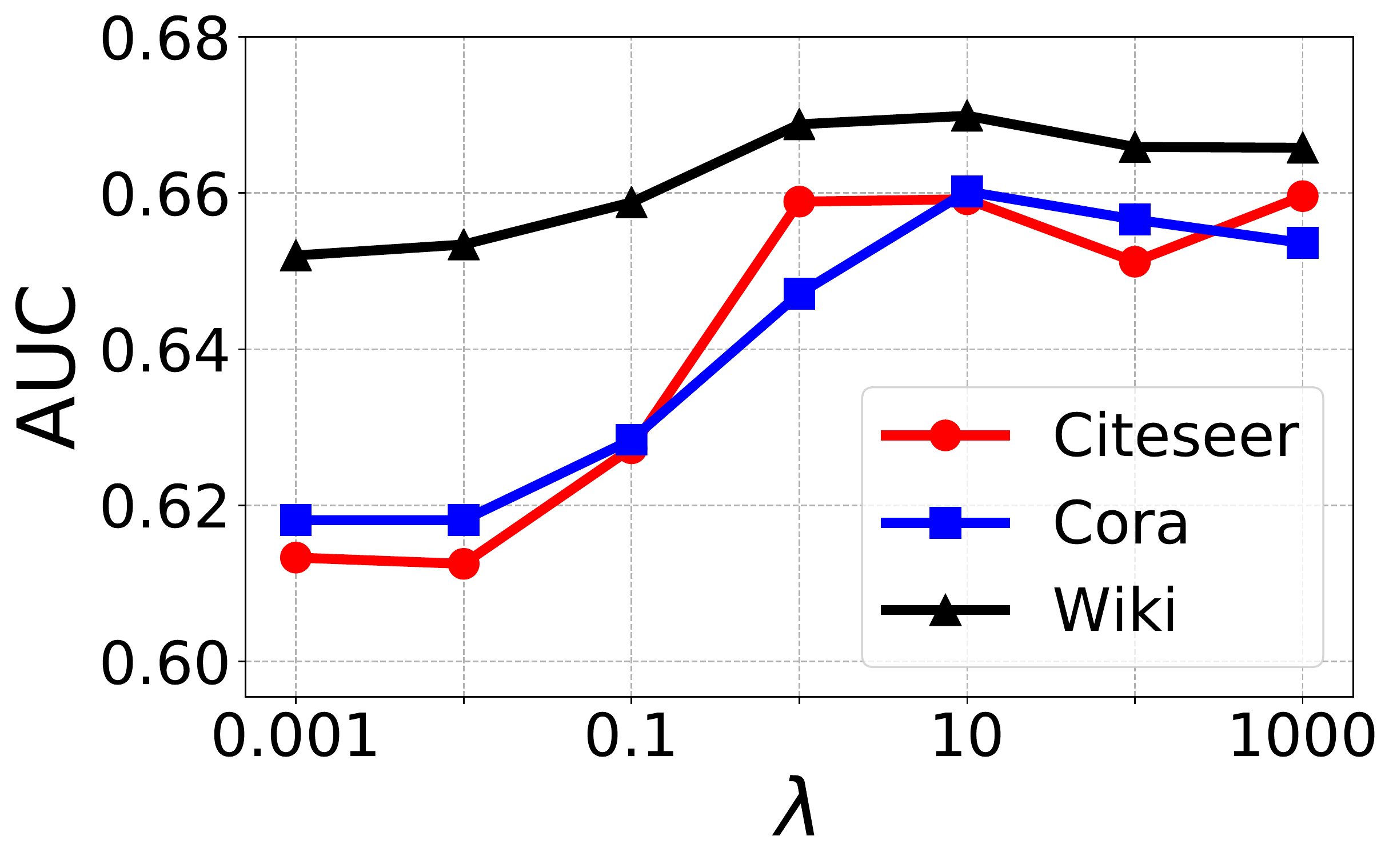}
	}
	\caption{Impact of hyperparameters on node classification (left, training ratio 50\%) and link prediction (right).}
	\label{fig:model-sensitivity}
\end{figure}

 Fig.~\ref{fig:eps} presents the experimental results when varying $\epsilon$ from 0.1 to 5.0. For both learning tasks, we can find that the performance in these three datasets first improves with the increase of $\epsilon$, and then drops dramatically after $\epsilon$ passing some threshold. It suggests that appropriate setting of $\epsilon$ improves the model robustness and generalization ability, while adversarial perturbation with too large norm constraint can destroy the learning process of embedding vectors. Besides, it can be easily noticed that the best settings of $\epsilon$ are different for different datasets in general. Specifically, Citeseer has the best results in both link prediction and node classification when $\epsilon=1.1$, Cora achieves the best results when $\epsilon=0.9$, while the best setting of $\epsilon$ for Wiki is around 0.5. Based on the experimental results on these three datasets only, it seems that the denser the network is, the smaller the best noise level parameter $\epsilon$ should be. 
 
 We conduct link prediction and node classification on three datasets with the adversarial regularization strength $\lambda$ from the set $\{0.001, 0.01, 0.1, 1, 10, 100, 1000\}$.
 Fig.~\ref{fig:reg_adv} displays the experimental results. For node classification, the best result is obtained when $\lambda$ is set to around 1, larger values can result in performance degradation. For example, the classification accuracy on Wiki drops dramatically when $\lambda$ reaches 10, and larger setting produces worse results. For link prediction, the performance is quite consistent among the three datasets. Specifically, when $\lambda$ increases from 0.001 to 10, the AUC score shows apparent increase for all datasets, and then tends to saturate or decrease slightly. Empirically, 1 is an appropriate value for the adversarial regularization strength $\lambda$.

\section{Related Work}\label{related_work}
 \textbf{Network Embedding}. Some early methods, such as IsoMap~\cite{Nature-00-Joshua} and LLE~\cite{Science-00-Sam}, assume the existence of a manifold structure on input vectors to compute low-dimensional embeddings, but suffer from the expensive computation and their inability in capturing highly non-linear structural information of networks. More recently, some negative sampling approach based models have been proposed, including \textsc{DeepWalk}~\cite{KDD-14-Bryan}, LINE~\cite{WWW-15-Jian} and node2vec~\cite{KDD-16-Grover}, which enjoys two attractive strengths: firstly, they can effectively capture high-order proximities of networks; secondly, they can scale to the widely existed large networks. \textsc{DeepWalk} obtains node sequences with truncated random walk, and learns node embeddings with Skip-gram model~\cite{NIPS-13-Tomas} by regarding node sequences as sentences. node2vec differs from \textsc{DeepWalk} by proposing more flexible random walk method for sampling node sequences. LINE defines first-order and second-order proximities in network, and resorts to negative sampling for capturing them.
 
 Further, some works~\cite{CIKM-15-SsCao,KDD-16-Mingdong,AAAI-17-XiaoW} tried to preserve various network structural properties in embedding vectors based on matrix factorization technique. GraRep~\cite{CIKM-15-SsCao} can preserve different $k$-step proximities between nodes independently, HOPE~\cite{KDD-16-Mingdong} aims to capture asymmetric transitivity property in node embeddings, while N-NMF~\cite{AAAI-17-XiaoW} learns community structure preserving embedding vectors by building upon the modularity based community detection model~\cite{PhysRevE.74.036104}. Meanwhile, deep learning embedding models~\cite{AAAI-16-SsCao,KDD-16-DxW,asunam-ShenC17,corr-abs-1901-01718} have also been proposed to capture highly non-linear structure. DNGR~\cite{AAAI-16-SsCao} takes advantages of deep denoising autoencoder for learning compact node embeddings, which can also improve model robustness. SDNE~\cite{KDD-16-DxW} modifies the framework of stacked autoencoder to learn both first-order and second-order proximities simultaneously. DNE-SBP~\cite{corr-abs-1901-01718} utilizes a semi-supervised SAE to preserve the structural balance property of the signed networks. Both GraphGAN~\cite{AAAI-18-Hongwei} and A-RNE~\cite{pakdd-qydqlaz19} leverage generative adversarial networks to facilitate network embedding, with the former unifies the generative models and discriminative models of network embedding to boost the performance while the latter focuses on sampling high-quality negative nodes to achieve better similariy ranking among node pairs.
 
 However, the above mentioned models mainly focus on learning different network structures and properties, while neglecting the existence of noisy information in real-world networks and the overfitting issue in embedding learning process. Most recently, some methods, including ANE~\cite{AAAI-18-Quanyu} and \textsc{NetRA}~\cite{KDD-YuZCASZCW18}, try to regularize the embedding learning process for improving model robustness and generalization ability based on generative adversarial networks (GANs). They have very complicated frameworks and suffer from the well-recognized hard training problems of GANs. Furthermore, these two methods both encourage the global smoothness of the embedding distribution, while in this paper we utilize a more succinct and effective local regularization method.
 
 \textbf{Adversarial Machine Learning}. It was found that several machine learning models, including both deep neural network and shallow classifiers such as logistic regression, are vulnerable to examples with imperceptibly small designed perturbations, called adversarial examples~\cite{ICLR-2014-Szegedy,goodfellow2014explaining}. This phenomenon was firstly observed in areas like computer vision with continuous input vectors. To improve model robustness and generalization ability, adversarial training method~\cite{goodfellow2014explaining} is shown to be effective. It generates adversarial perturbations for original clean input with the aim of maximizing current model loss, and further approximates the difficult optimization objective with first-order Taylor Series. Such method has also been applied to text classification problem in~\cite{ICLR-16-Miyato,IJCAI-SatoSS018} by defining the perturbation on continuous word embeddings, and recommendation in~\cite{SIGIR-2018-xiangnan} by generating adversarial perturbations on model parameters. However, to the best of our knowledge, there is no practice of adversarial training regularization for graph representation learning. 
 
 For graph structured data, they are fundamentally different from images because of their discrete and indifferentiable characteristics. Some existing works~\cite{ICML-DaiLTHWZS18,KDD-ZugnerAG18,corr-abs-1809-02797} aimed to explore how to generate the adversarial examples in the discrete, binary graph domain, and whether similar vulnerability exists in graph analysis applications. In~\cite{ICML-DaiLTHWZS18}, adversarial attacks are generated by modifying combinatorial structure of graph with a reinforcement learning based method, which is shown to be effective in Graph Neural Network models. Both~\cite{KDD-ZugnerAG18} and~\cite{corr-abs-1809-02797} designed attack methods to Graph Convolutional Network~\cite{ICLR-16-KipfW}. Particularly, NETTACK~\cite{KDD-ZugnerAG18} focuses on attributed graph classification problem and FGA~\cite{corr-abs-1809-02797} tackles network representation learning. However, all of them studied adversarial attack methods without providing any defense algorithms for improving the robustness of existing methods against these attacks. Differently, in this paper, we aim to propose adversarial regularization method for network embedding algorithms to improve both model robustness and generalization ability.

\section{Conclusion} \label{conclusion}
 In this paper, we proposed two adversarial training regularization methods for network embedding models to improve the robustness and generalization ability. Specifically, the first method is adapted from the classic adversarial training method by defining the perturbation in the embedding space with adaptive $L_2$ norm constraint. Though it is effective as a regularizer, the lack of interpretability may hinder its adoption in some real-world applications. To tackle this problem, we further proposed an interpretable adversarial training method by restricting the perturbation directions to embedding vectors of other nodes, such that the crafted adversarial examples can be reconstructed in the discrete graph domain. Both methods can be applied to the existing embedding models with node embeddings as model parameters, and \textsc{DeepWalk} is used as the base model in the paper for illustration. Extensive experiments prove the effectiveness of the proposed adversarial regularization methods for improving model robustness and generalization ability. Future works would include applying adversarial training method to the parameterized network embedding methods such as deep learning embedding models.

\section{Acknowledgments}
 Parts of the work were supported by HK ITF UIM/363.

\bibliographystyle{ACM-Reference-Format}
\balance 
\bibliography{main}

\end{document}